\documentclass[1p,sort&compress]{elsarticle}
\usepackage{amssymb}
\usepackage{graphicx}
\usepackage{subfigure}
\usepackage{hhline}
\usepackage{bm}
\usepackage{booktabs}
\usepackage{epstopdf}
\usepackage{url}
\usepackage[ruled,vlined]{algorithm2e}
\usepackage{slashbox}
 \usepackage{lineno}
 \usepackage{setspace}
 \usepackage[latin1]{inputenc}
\begin{document}
\doublespacing
\begin{frontmatter}

\title{Scalable Prototype Selection by Genetic Algorithms and Hashing~\tnoteref{t1}}
\tnotetext[t1]{Parts of the work described have been published as part of the first author's PhD Thesis~\cite{Plasenciaphd2015}.}

\author[lab1,lab3]{Yenisel Plasencia-Cala\~{n}a\corref{cor1}}
 \ead{yplasencia@cenatav.co.cu}
 \author[lab2]{Mauricio Orozco-Alzate}
 \author[lab1]{Heydi M\'{e}ndez-V\'{a}zquez}
 \author[lab1]{Edel Garc\'{i}a-Reyes}
 \author[lab3]{Robert P.W.~Duin}


\address[lab1]{Departamento de Reconocimiento de Patrones, Centro de Aplicaciones de Tecnolog\'{i}as de Avanzada, CENATAV. 7a A \#21406 esq. 216, Rpto.~Siboney, Playa, C.P.~12200. Ciudad de la Habana, Cuba}
\address[lab2]{Universidad Nacional de Colombia - Sede Manizales - Departamento de Informática y Computación - km 7 vía al Magdalena, Manizales 170003 - Colombia}
\address[lab3]{Pattern Recognition Laboratory, Delft University of Technology. P.O. Box 5031, 2600 GA, Delft, The Netherlands}
\cortext[cor1]{Corresponding author. Tel.: +53 7 2720422; fax: +53 7 2724613.}
%
%
%

%
%

\begin{abstract}
Classification in the dissimilarity space has become a very active research area since it provides a possibility to learn from data given in the form of pairwise non-metric dissimilarities, which otherwise would be difficult to cope with. The selection of prototypes is a key step for the further creation of the space. However, despite previous efforts to find good prototypes, how to select the best representation set remains an open issue. In this paper we proposed scalable methods to select the set of prototypes out of very large datasets. The methods are based on genetic algorithms, dissimilarity-based hashing, and two different unsupervised and supervised scalable criteria. The unsupervised criterion is based on the Minimum Spanning Tree of the graph created by the prototypes as nodes and the dissimilarities as edges. The supervised criterion is based on counting matching labels of objects and their closest prototypes. The suitability of these type of algorithms is analyzed for the specific case of dissimilarity representations. The experimental results showed that the methods select good prototypes taking advantage of the large datasets, and they do so at low runtimes.
\end{abstract}


\end{frontmatter}
\section{Introduction}
The vector space representation is a common option to represent the data for learning tasks since many statistical techniques are applicable for this kind of representation. However, there is an increasing number of real-world problems which are not vectorial. Instead, the data are given in terms of pairwise dissimilarities which may be non-Euclidean and even non-metric. Dissimilarity-based representations are becoming more and more useful for real world applications, for example we found heterogeneous real world data where the variables are of different nature and a dissimilarity measures can be designed to cope with this problem. There is also the example of metric learning methods, which are becoming the standard for comparing vectorial representations, due to their ability to create discriminative metrics trained specifically for each problem. In~\cite{pekalskaDissbook} several approaches were presented to learn from dissimilarity data, where the dissimilarity space (DS) has several advantages over the other approaches. In the DS approach, the dissimilarities of the training objects to the representation set are interpreted as coordinates of a vector space.\\
\indent A careful selection of the prototypes is needed to build a ``good" DS based on a small set of prototypes. However, a random selection was found to perform well for large numbers of prototypes~\cite{Pekalskaprotsel2006}. Since this is the fastest method, the selection of good prototypes by more dedicated methods is only of interest for small numbers of prototypes, i.e. if there are constraints in storage and speed for the deployments of the procedures. A good prototype selection method must be able to find a minimal set without a significant decrease in the accuracy of classifiers in the DS.\\
\indent Several methods have been proposed~\cite{Pekalskaprotsel2006,Bunke2008} to find small representation sets by supervised or unsupervised strategies. Supervised methods have the advantage of maintaining a high accuracy, however this is achieved at high computational costs; besides, they might suffer from overfitting. Unsupervised methods have the advantages of being fast and avoiding overfitting. However, as a disadvantage, they are not always good in maintaining class separability since class labels are not taken into account.\\
In this paper we focus on the selection of prototypes for classification in the DS. Our more specific target is to learn from large datasets, thereby we must design scalable procedures since it is known that the full search for the best prototypes set is intractable. This has been overlooked so far and only some studies such as the one in~\cite{Olivetti2012} cope with the problem. The analysis of large datasets is of interest because, nowadays, vast amount of data arise due to dropping costs for capturing, transmitting, processing and storing. There are also modalities that have millions of classes such as biometrics and others that present hundreds of thousands samples such as brain tractography data in~\cite{Olivetti2012}.\\
\indent In~\cite{plasencia2014}, we proposed prototype selection methods based on genetic algorithms (GAs), but these procedures were only applicable to datasets where the full dissimilarity matrix can be loaded in memory. Therefore, the procedures cannot cope with very large datasets with hundreds of thousands of objects since they do not fit into memory, in addition they did not scale-up for datasets in the order of millions. Following up on~\cite{plasencia2014}, we create new versions of the proposed procedures which are able to cope with the case when the full dissimilarity matrix does not fit into memory and to scale up to larger datasets in terms also of speed. The main improvements in the new procedures are: 1) we propose to perform the computations of dissimilarities on demand inside the GA procedure; 2) we proposed to use Spherical Hashing~\cite{Lee:2012:SH} to speed up the computations. The proposed GAs also perform a fast clustering replacing the random initialization of the standard GAs which allow to converge faster to a good solution. Here we shortly recapitulate the theory related to our proposals. We present the modified GAs and their complexity analysis. A set of new experiments is performed to validate our proposals using real world data containing hundreds of thousands of objects. \\
\indent The selection of prototypes seems similar to the selection of features for feature spaces. However, the interpretation of features is different from the interpretation of prototypes since features might be very different and unique while dissimilarities relate similar objects. Therefore, adequate methods for selecting features are not necessarily adequate to select prototypes. A set of prototypes as well as vectors of dissimilarities are homogeneous in the sense that they represent values of the same dissimilarity measure; however a set of features may be very different since features may correspond to different type of measurements, even not numerical e.g. categorical. Thereby, the comparison of features is ill-defined. In contrast, the comparison of objects is well defined and it can be performed in a natural manner by an expert-defined dissimilarity measure or by an appropriate distance measure. In large datasets, objects have similar neighbors that may cause (after replacement) just slightly better or worse results. These are convenient properties that encourage the use of GAs.\\
\indent It is usually assumed in the literature that linear-time algorithms are acceptable for scaling up to large datasets~\cite{scalePedrajas12}. We also adopt this assumption. The parameter that dominates the complexity in our problem is the number of samples, since we assume that the other parameter, the number of selected prototypes, will always be small. Some strategies to cope with scalability~\cite{scalePedrajas12} include parallelism, techniques to work with data that do not fit into memory, stochastic methods, etc. In our new proposal, since dissimilarities are computed on demand, the space complexity is smaller than in the first version of the proposed methods but at the cost of increased time complexity, which is a function of dissimilarity computation complexity. \\
The remaining part of the paper is organized as follows: Section~\ref{sec:ds} presents the dissimilarity representation and prototype selection, Section~\ref{sec:methods2} presents the two proposed methods and their complexity analysis considering the computation of dissimilarities on demand. Experimental results are reported and discussed in Section~\ref{sec:results} and conclusions are drawn in Section~\ref{sec:concl}.
\section{Dissimilarity Space and Prototype Selection}
\label{sec:ds}
\indent The dissimilarity space was proposed by Pekalska and Duin~\cite{pekalskaDissbook}. It was postulated as a Euclidean space, which allows the use of all the classifiers that assume such space. Let $X$ be the space of objects which may not be a vector space, but an input space of row measurements. Let $R =\{r_{1},r_{2},...,r_{k}\}$ be the set of prototypes such that $R\in X$, and let $d:X\times X\rightarrow{\mathbb{R}^{+}}$ be a suitable dissimilarity measure for the problem. The prototypes may be chosen based on some criterion or even at random; however, the goal is that they have good representation capabilities specially when pursuing small representation sets. For a finite training set $T =\{x_1,x_2,...,x_l\}$ such that $T\in X$, the dissimilarity space is created by the data dependent mapping $\phi^{d}_{R}:X \rightarrow {\mathbb{R}}^{k}$ where:
\begin{equation} \label{eq:dis}
  \phi^{d}_{R}(x_i) = [d(x_i,r_{1}) \hspace{1.5mm} d(x_i,r_{2})\hspace{1.5mm}...\hspace{1.5mm}d(x_i,r_{k})].
\end{equation}
\indent Many methods have been proposed for the selection of prototypes in small-sized datasets. The study in~\cite{Pekalskaprotsel2006} presents the Kcentres cluster-based method, an editing and condensing method, as well as supervised methods such as the forward selection (FS) to select prototypes. However, only the Kcentres is able to scale to large datasets. The methods proposed in~\cite{stringprototypes,Bunke2008} to select prototypes for strings and graph problems maintain in general a linear computational complexity. However, only two of the methods have good performances: the Kcentres and a version of the farthest first transversal (FFT) which they call spanning prototype selector. Thereby, we include them in our comparisons. We assume we have a validation set $V$ which is used to select the prototypes out of it. The validation set is also used to compute those selection criteria that involve dissimilarity computations between the prototypes and objects in the dataset. In very large datasets we can afford large validation sets, but at the same time this poses the challenge to create methods which are able to deal with such large sets. It is worth noting that Kcentres has a complexity of $O(nk+k*(n/k)^2)$ and FFT has a complexity of $O(nk)$, where $k=|R|$ and $n=|V|$. For the sake of simplicity we say that they both behave as linear or almost linear time algorithms. The method proposed in~\cite{Olivetti2012} presents a variant of FFT that uses a sample of the dataset.  However, the sampling decreases performance of the original version. Thereby, we use the version computed on the full dataset for our experiments.
\section{Scalable Genetic Algorithms}
\label{sec:methods}
\indent Here we recapitulate the theory related to the improved GAs based on the previously proposed ones   in~\cite{plasencia2014}. We propose two slightly different variants of GAs for prototype selection which are designed in order to be able to cope with large datasets that do not fit into memory. The two methods receive as parameter the desired number of prototypes. Finding an appropriate number of prototypes for each particular problem is of interest since it allows one to save computing time and execute prototype selection methods only for a particular cardinality of the set. A good practice is to find the intrinsic dimensionality and select the number of prototypes accordingly. In our setup, we do not have a square dissimilarity matrix $D$ computed among all training samples. The prototypes will be selected from the set of samples. Dissimilarities must be measured with these prototypes on demand on each iteration of the method.\\
\indent The GA is a search method based on heuristics that mimic the natural evolution mechanisms, by evolving individuals (chromosomes or solutions) created after each generation by the best fitted ones. In our problem, each individual is a set of prototypes of fixed cardinality $k$ codified in a $k-vector$ containing in each position the index of the potential prototype. For example, the $5-vector$ $(65,30,7,19,87)$ codifies an individual representing a set of 5 potential prototypes which can be accessed in some data structure by the indexes 65, 30 and so on. The GA usually starts the search in an initial population of randomly generated individuals. \\
\renewcommand{\floatpagefraction}{0.75}
\incmargin{1em}
\linesnumbered
\begin{algorithm}[h!]
  \restylealgo{boxed}
  \SetLine
  \KwIn{$F$: feature matrix with $n$ objects in rows and $m$ features in columns;  $k$: desired number of prototypes, $S$: number of individuals in the population, $rp$: reproduction probability, $mp$: mutation probability, $iter$: number of generations}
  \KwOut{bestindividual: set of prototypes indexes }
  \SetKwFunction{GenerateInitialP}{GenerateInitialPopulation}
  \SetKwFunction{Fitn}{Fitness}
  \SetKwFunction{compdiss}{ComputeDissimilarities}
   \SetKwFunction{clustera}{Nearest prototype clustering assignment}
  \SetKwFunction{Reprod}{Reproduce}
  \SetKwFunction{Mut}{Mutate}
  \SetLine
    $D  \leftarrow \compdiss{$F(1..n,1...m),F(w,1...m)$}$\;
  \tcp{perform a nearest prototype clustering assignment to randomly chosen centers in the space of candidates to prototypes to find $k$ clusters}
    $cluslabs \leftarrow$ \clustera{$D,k$}\;
  \tcp{randomly generate the population ensuring that, in the $j$-th position of the individual, only objects belonging to the $j$-th cluster are allowed}
    $P \leftarrow$ \GenerateInitialP{$cluslabs,F,k,S$}\;
        $bestindividual \leftarrow P[1]$\;
\tcp{find the best solution from the population  and assign it to bestindividual}
              \ForEach{currentindividual in P}
         {
         \tcp{Note that the next line is where the proposed selection criteria must be used as the Fitness function}
             \If{\Fitn{$currentindividual,F$} $>$ \Fitn{$bestindividual,F$}}
             {
              $bestindividual \leftarrow currentindividual$\;
             }
         }
    \While{number of generations $<$ $iter$}
     {
      \tcp{Evolution cycle}
       \ForEach{currentindividual in P}
         {
          \tcp{Reproduction, replace a gene of currentindividual with probability $rp$ by a gene of the best solution}
              \Reprod{$bestindividual,currentindividual,rp$}\;
               \tcp{Mutation, change a gene of currentindividual with probability $mp$}
              \Mut{$currentindividual,mp$}\;
          }
          \tcp{find the best solution from the population  and assign it to bestindividual}
       }
  \caption{Scalable Genetic Algorithm \label{alg:GA}}
  \end{algorithm}
\decmargin{1em}
\indent Before executing the GA, we propose an initialization for the method, which performs a nearest prototype clustering assignment to randomly chosen centers in the set of candidates to prototypes to find $k$ clusters where the number of clusters equals the desired number of prototypes. The candidates are clustered in order to guide the GA search in such a way that it has a faster convergence. The clustering runtime is $O(nk)$, where $n=|V|$ and $k=|R|$. The GA is slightly modified since its initial population is now generated by more restricted chromosomes. Each prototype represented in a position or gene of a chromosome is randomly sampled from a different cluster. In each generation, the best solution according to the fitness function is found and reproduced with each member of the population with a preset probability for the genes using uniform reproduction or crossover. Elitist selection is performed since the best fitted individual is retained for the next generation without undergoing mutation. In addition, only the best fitted individual is selected as parent of the next population of individuals. The rest of the population undergoes gene mutation with a preset probability which is usually small. We maintain the constraint that the new index codified in a gene must belong only to the specific cluster linked to the gene. \\
\indent In order to be fast and achieve full scalability, these methods should be able to handle: (1) large sets of candidates for prototypes, (2) large numbers of individuals in the search space of the GA and (3) large number of samples ($|V|$) to be used (if the selection criterion requires it) to compute the fitness function. In our proposal the GA handles well (1) large sets of candidates for prototypes since we do not use the standard binary codification of individuals that demands vectors of length equal to $n$ where $n\gg k$. Instead, we resorted to vectors of length equal to the number of prototypes $k$ since we codify only the indexes of the prototypes to be evaluated. Scalability in the number of individuals to analyze in the search space (2) is also achieved since the stopping condition is a small predefined number of GA generations that does not depend on the number of individuals in the search space. A small number of generations is sufficient for GA's convergence thanks to its guided sampling since not all the possible combinations of prototypes are explored but only the best ones which arise after each generation. In addition, if the initial clustering is used, it helps to avoid redundant prototypes in the same individual. Scalability in the number of samples to be used in the computation of the fitness function (3) will be explained in the subsections explaining the proposed criteria.
\section{Proposed Genetic Algorithms When Dissimilarities Must Be Computed On Demand}
\label{sec:methods2}
In this section we assume that the full dissimilarity matrix cannot be loaded into memory and we perform our analysis assuming that the dissimilarities have not been pre-computed and stored. The dissimilarities must be computed on demand using a proper dissimilarity measure for the problem. In addition, the data is given by raw measurements or some other intermediate representation, e.g. images/time signals/graphs and we assume that finding a dissimilarity between any two objects needs $q$ computations. The proposed GAs are modified accordingly allowing scalability for such large datasets where the storing of the full dissimilarity matrix into memory or even on disk may not be feasible or desirable. In this section we will describe the main components of the proposed GAs that change in this new situation as well as their computational complexity analysis. We assume that the complexity of the dissimilarity measure is linear in the number of measurements. This is true, for example, for distances such as those of the Minkowski family which contains the widely used Euclidean one.\\
\indent First, in the clustering, the nearest prototype assignment to random centers must include the computation of dissimilarities among the randomly initialized prototypes and all the samples, but this is performed only once so the total cost is $O(nkq)$, being $q$ the number of measurements. The two important links of a general purpose GA with an specific problem are the encoding of solutions or individuals and the fitness function. The encoding of solutions is affected by the need to compute dissimilarities on demand since we access now to the object instead of its already pre-computed dissimilarities with other objects. The pseudo-code is presented in Algorithm~\ref{alg:GA}. In the next subsections the theory related to the proposed fitness functions and their modifications are explained.
\subsection{Minimum Spanning Tree-Based Unsupervised Criterion}
\label{sec:MST}
\indent In the fitness function computation, the set of $k$ prototypes being evaluated as well as the $n$ validation samples are involved. The proper number of prototypes $k$ depends on the intrinsic dimension of the data which is usually small for real world problems, thereby, $n\gg k$. For large datasets, this implies that the dominant term for the fitness computation is the total number of samples $n$. To achieve scalability in the fitness function, it must be able to scale to a large $n$. This highly depends on how the criterion to be optimized in the fitness function by the GA is conceived. Our first proposal for GA criterion is based on the minimum spanning tree (MST) of a set of prototypes. Prototypes are interpreted as nodes in a graph and dissimilarity values between prototypes correspond to edge weights. The sum of edge weights (usually named tree weight) is used as criterion to be maximized, thereby increasing the diversity of the prototypes and improving the coverage over the DS. \\
\indent The MST weight is related to the R\'{e}nyi entropy of the set of prototypes~\cite{MSTimages2000}. This relation is monotonically increasing: the entropy of the set of prototypes increases as the MST weight increases. The entropy is also a type of diversity measure which confirms our intuition that higher MST weights are related to higher diversity of the prototypes.\\
\indent As we used the Prim's algorithm to find the MST and the graph of dissimilarities among the prototypes is complete, the computation of this criterion from an already constructed graph has a runtime of $O(k^2\log(k))$. Therefore, it is independent on the large number of samples $n$ and, as a consequence, highly scalable for very large problems. The modification that we introduce in the unsupervised fitness requires to compute the all vs. all pairwise dissimilarities among the candidate prototypes indexed in an individual. The second step is to compute the fitness value which includes the MST construction. The pseudo-code is presented in Algorithm~\ref{alg:unsupcrit}.\\
\indent The total fitness function, including first the computation of the square pairwise dissimilarity matrix for the prototypes and second the MST computation, takes $O(k^{2}q+k^2\log(k))$, which boils down to $O(k^{2}(q+\log(k)))$. In case the clustering step is used, the total runtime of the unsupervised GA is $O(nkq + k^{2}(q+\log(k)))$. If the clustering step is not performed, the GA method takes $O(k^{2}(q+\log(k)))$. \\
\incmargin{1em}
\linesnumbered
\begin{algorithm}[h!]
  \restylealgo{boxed}
  \SetLine
  \KwIn{$w$: vector of prototypes indexes; $F$: feature matrix with $n$ objects in rows and $m$ features in columns}
  \KwOut{$fitnessvalue$: fitness value }
  \SetKwFunction{Prim}{Prim}
  \SetKwFunction{compdiss}{ComputeDissimilarities}
  \SetKwFunction{SumWeights}{SumWeights}
  \SetLine
  $D  \leftarrow \compdiss{$F(w,1...m),F(w,1...m)$}$\;
  \tcp{Interpret the prototypes indexed in $w$ as nodes and dissimilarities among them stored in D as edges weights of a complete graph G = (v,e)}
    \tcp{compute minimum spanning tree by Prim's algorithm}
    $v' \leftarrow v[1]$\;
    $k \leftarrow |v|$\;
    $e' \leftarrow \emptyset$\;
    \While{$|e'| < k-1$}
     {
     \tcp{select an edge of minimum weight which connects one node in v' with a node which is not in v'}
     \tcp{add the new edge to e', add the new node to v'}
     }
     $MST \leftarrow (v',e')$\;
     \tcp{sum all the weights of edges $e'$ in MST}
     $fitnessvalue \leftarrow$ \SumWeights{MST}\;
  \caption{Unsupervised Fitness function by MST \label{alg:unsupcrit}}
\end{algorithm}
\decmargin{1em}
\subsection{Supervised Criterion Based on Counting Matching Labels}
\indent Our second proposed criterion is a linear-time supervised criterion that is different from previous supervised ones~\cite{ZareBorzeshi2013} for prototype selection, since it does not compute a classification error in the DS or an intra-class distance, which are usually quadratic. Our method, instead, considers each candidate for prototype as a representative of a cluster and every object in $V$ is assigned to its nearest cluster represented by the prototype. The proposed criterion counts the number of assigned objects whose class labels match their nearest protototype class label. The best solution is the one that maximizes this value. The runtime complexity is ($O(nk)$).\\
\indent The supervised fitness function is reformulated when the dissimilarities between the samples are not given. In this case, the dissimilarities between all the objects and the prototypes must be computed before the criterion. The pseudo-code including this modification is presented in Algorithm~\ref{alg:supcrit}. The total complexity of the fitness function, including the computation of dissimilarities between objects and prototypes, is now $O(nkq)$. The total runtime remains the same whether the clustering step is used or not. The computational complexities for both cases, when the dissimilarities are given as in~\cite{plasencia2014}, or when they must be computed on demand are summarized in Table~\ref{ccomp}.
\subsection{Spherical Hashing to speed-up the criterion computation}
\indent As the proposed supervised criterion uses all the data in the validation set to compute the fitness value for each solution, it would be convenient to speed-up its computation especially for data in the order of millions. In the unsupervised proposal only a small amount of data is used in each iteration to compute the criterion but it might also benefit from a faster computation. We propose a solution for this which is able to work with dissimilarity data. The method is  based on incorporating in the fitness function computation a hashing method proposed in the context of approximate nearest neighbour search. The approaches for approximate nearest neighbour search are mostly based on tree structures~\cite{muja2009fast}, hashing functions~\cite{andoni2006near} or Product Quantization\cite{Jegou:2011:PQN}. We analyze these methods to find those able to work with dissimilarities directly. The analysis showed that from the trees family the Vantage Point Tree~\cite{Yianilos:1993:VPT} is able to fulfill this condition, but it has to compute an all vs all dissimilarity matrix which is prohibitive for large datasets. From the hashing family the Spherical Hashing~\cite{Lee:2012:SH} was also able to use the dissimilarity data directly. Hashing functions are able to create compact codes which allow to speed-up the search in large datasets. The Spherical Hashing method computes a set of pivots and each dataset object is codified by its dissimilarity to the pivots. The cardinality of the set of pivots is usually a power of 2 for convenience, e,g, 64, 128 or 256. Each pivot has a specified trained radius, and a binary code is assigned by thresholding and assigning 1 if the distance to the pivot is smaller than threshold and 0 otherwise. We found the base codification by dissimilarities more suitable for our work instead of the binary codification since it provides more relevant information than the binary codes while also speeding-up the process. This can also be seen as an intermediate dissimilarity representation where the prototypes are randomly selected, which in our case is used for selecting better prototypes. 

\incmargin{1em}
\linesnumbered
\begin{algorithm}[h!]
  \restylealgo{boxed}
  \SetLine
  \KwIn{$w$: vector of prototypes indexes; $F$: feature matrix with $n$ objects in rows and $m$ features in columns}
  \KwOut{$fitnessvalue$: fitness value }
  \SetKwFunction{argmin}{argmin}
    \SetKwFunction{compdiss}{ComputeDissimilarities}
  \SetKwFunction{getclasslabel}{getclasslabel}
  \SetLine
  \tcp{interpret the prototypes $r_j$ indexed in $w$ as centers of clusters and compute dissimilarities of all the $x\in T$ to the $r_j$}
  $D  \leftarrow \compdiss{$F(1...n,1...m),F(w,1...m)$}$\;
  $fitnessvalue \leftarrow 0$\;
  \ForEach{$x\in T$}
         {

         \tcp{find the nearest prototype of $x$}
         $r' \leftarrow$ \argmin{$D[x,r_j]$}\;
        \If{\getclasslabel{x}=\getclasslabel{r'}}
             {
             $fitnessvalue \leftarrow fitnessvalue+1$\;
             }
         }
  \caption{Supervised Fitness function based on counting matching labels \label{alg:supcrit}}
\end{algorithm}
\decmargin{1em}

\begin{table*} [!ht] \centering
\caption{Computational complexities when dissimilarities are given and when dissimilarities are computed on demand}
 \label{ccomp}\small
\begin{tabular}[c]{|c|c|c|c|c|}
\hline
Type of problem& Fitness function & GA with clust. & GA no clust.\\
 \hline
 Given diss. unsup.& $O(k^2\log(k))$&$O(nk+k^2\log(k))$& $O(k^2\log(k))$\\
\hline
 Given diss. sup.& $O(nk)$&$O(nk)$& $O(nk)$\\
\hline
On demand diss. unsup.& $O(k^{2}(q+\log(k)))$&$O(nkq+ k^{2}(q+\log(k)))$& $O(k^{2}(q+\log(k)))$\\
\hline
On demand diss. sup.& $O(nkq)$&$O(nkq)$& $O(nkq)$\\
\hline
\end{tabular}
\end{table*}\normalsize
\section{Experiments}
\label{sec:results}
\subsection{Datasets and Experimental Setup}
\begin{figure}[!ht]
\centering
\subfigure[MNIST dataset]{
\includegraphics[scale=0.3]{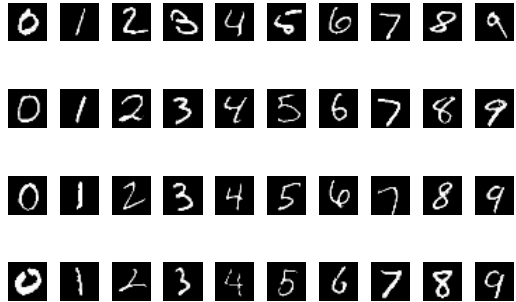}
\label{digMNIST}
}
\subfigure[SVHN1 dataset]{
\includegraphics[scale=0.3]{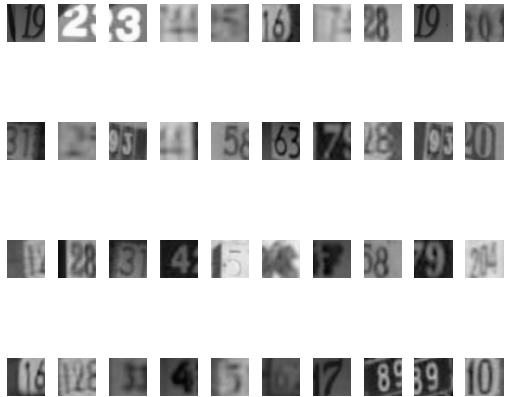}
\label{digSVHN}
}
\caption{Examples of grayscale images from the datasets of digits used in the experiments}
\label{digits}
\end{figure}

\indent Four different datasets~\cite{plasencia2014} of small to medium size were used for the experiments considering that the full dissimilarity matrix fits into memory. Some results computed on them are presented here to analyze their performance in terms of classification accuracies. \\
\indent Another four datasets of large size were used here for the experiments to test the classification accuracies but mainly the scalability of the proposed methods for large datasets where the full dissimilarity matrix does not fit into memory. \\
\emph{MNIST data.} The dataset~\cite{lecun98} was collected from subsets of NIST having a balanced number of digits written by high-school students and Census Bureau employees. The original black and white images from NIST were scaled to fit in a 20x20 pixel box while preserving their aspect ratio. The digits were centered in a 28x28 image by computing the center of mass of the digit image pixels, and translating the digit image to fit the center in that of the 28x28 image. Our dissimilarities were created using Euclidean distances on the $28\times 28$ images. \\
\emph{Street View House Numbers data.} The dataset is used for the purpose of digit recognition in the wild~\cite{SVHNpaper}. We divided it into SVHN1 and SVHN2 depending on which of the available sets (the smaller with 73 257 or the larger with 531131 images) is used as validation set. Examples of digits from MNIST and this dataset are shown in Fig.~\ref{digits}. For SVHN1 and SVHN2 we had to process the $32\times 32$ digit images first since they are very noisy. The digit histograms were equalized and the resulting image intensities were scaled by a Gaussian-shaped function emphasizing the middle of the images which actually contain the digits and gradually giving less weight to farther pixels which contain noise. Next, the images were blurred using a $5\times5$ Gaussian kernel with $\sigma$ parameter equal to 0.5. SVHN1 is the smaller dataset with 73 257 images containing the most difficult samples. SVHN2 is larger, with 531 131 extra data containing easier digits. These sets are later partitioned into two sets, one for prototype selection and one for training. The standard test set of 26 032 images is used for testing.\\
\emph{YouTube Faces dataset.} The original version of YouTube faces database~\cite{Youtubeface} is composed by face videos and it was designed to investigate unconstrained face recognition. It contains a total of 3425 videos from YouTube of 1595 subjects. The shortest clip contains 48 frames, the longest clip contains 6070 frames, while the average length of a video clip is 181.3 frames. In the database, there are 1045 subjects with at least one video having more than 100 frames. We used these subjects for our experiment, and only one video was selected for each person. The videos were split into frames to construct the datasets. Local Binary Patterns (LBP) descriptors were computed for the normalized faces contained in the frames and Euclidean distances are used as dissimilarity measure, since they behave well on these descriptors. Besides, these Euclidean distances are in agreement with our computational complexity analysis since we assume the use of a linear time dissimilarity measure. The characteristics of the datasets as well as the cardinality of the training sets used are summarized in Table~\ref{comp}.\\
\begin{table*} [!ht] \centering
\caption{Characteristics of the datasets used in this study, the last column ($|V|$) refers to the validation set cardinality used for the experiments}
 \label{comp}\small
\begin{tabular}[c]{|c|c|c|c|c|c|}
\hline
 Datasets & \# Classes & \# Obj& Metric &$|V|$\\
 \hline
Zongker & 10 & $200\times10$ & no& 1000\\
\hline
Diabetes & 2 & $500/268$ & yes & 384\\
\hline
Pendigits & 10 & 10992 & no & 5000\\
\hline
XM2VTS & 295 & $12\times295$ & yes& 1770\\
\hline
MNIST & 10 & $70 000$ & yes& 60 000\\
\hline
SVHN1 & 10 & $99 289$ & yes& 73 257\\
\hline
SVHN2 & 10 & $630 420$ & yes& 531 131\\
\hline
YouTube & 1045 & $308 963$ & yes& 247 170\\
\hline
\end{tabular}
\end{table*}\normalsize
\indent The small and medium-sized datasets were divided into validation, training and test set 30 times. The validation is used to select the prototypes out of it and to compute the selection criterion. In the case of MNIST, the standard training and test set division was used, except that the training set was randomly divided 10 times into one set for validation with 98\% of the total objects (without considering the test set), and the other 2\% was used to train the classifier. This was done since our purpose is to show that the selection methods scale well to large datasets and still find good prototypes. Small training sets were used since they give us the opportunity to analyze better if the selection with the proposed methods is successful since, in some cases, a large training set may compensate for an inadequate prototype selection. Similarly, in the case of SVHN1 we randomly divided 10 times the difficult set of 73 257 images to use 98\% for selecting the prototypes and 2\% to train the classifiers. We also carried out an experiment on SVHN2 where we selected the prototypes out of the extra set of SVHN containing 531131 images, we train with a subset of 53 113 images randomly selected and test with the standard test set, to show the scalability for half a million images. \\
\indent In the case of the YouTube dataset, we randomly divided the data into 80\% for validation from which the 0.05\% was used for training, and 20\% for testing. The best globally performing classifier per dataset between the linear discriminant classifier (LDC), which is the Bayes classifier assuming normal densities with identical covariance matrices, and the 1-NN classifier was used to report the classification errors for the different prototype selection methods compared in the medium-sized dataset, the compared methods are:
\begin{itemize}
\item Random selection
\item Forward selection~\cite{Pekalskaprotsel2006} optimizing the supervised criterion
\item FFT~\cite{Olivetti2012}
\item Kcentres~\cite{Pekalskaprotsel2006}
\item GA in the space of clustered prototypes with the proposed unsupervised fitness function based on MST (GA (clust) MST)
\item GA with the proposed unsupervised fitness function based on MST without clustering the prototypes (GA MST)
\item GA in the space of clustered prototypes with the proposed supervised fitness function (GA (clust) sup)
\item GA with the proposed supervised fitness function (GA sup)
\end{itemize}
In the case of the large datasets the classifiers tested were the 1-NN and the quadratic discriminant classifier (QDC) since we can afford to estimate different covariances for large datasets. In addition, the QDC outperforms the LDC for all the procedures on these datasets. The FS was not considered for the experiments on the large datasets due to its lack of scalability.\\
\indent The parameters used for the GA are: 20 individuals for the initial population, 0.5 for probability of reproduction per gene, and 0.02 for probability of mutation per gene. The stopping condition is 20 generations reached for the GA with initial clustering in the space of prototypes, and 25 for the GA without the clustering for the medium-sized datasets. These numbers are different in order to show that the clustering allows a faster convergence of the GA even when some advantage is given to the version without the clustering. For the large-sized datasets, both versions use 20 generations as stopping criterion and the probability of mutation is increased to 0.1 to allow a higher diversification of the solutions and therefore better exploration of the large search space.
\subsection{Results and Discussion}
\label{sec:results2}

\begin{table*} [!ht]\centering
\caption{Average classification errors and standard deviations for 20 prototypes in the medium size datasets. Best results are in bold.}
\label{resultssmall}\small
\begin{tabular}{|c|c|c|c|c|}
\hline
\backslashbox{Method}{\ Dataset}&Zongker(LDC)&Diabetes(LDC)&Pendigits(1-NN)&XM2VTS(LDC)\\
\hline
FFT           &$0.162\pm    0.01$&	$\bm{0.254\pm	0.01}$&	$0.016\pm	0.002$&	$0.428\pm	0.01$\\
\hline
GA sup        &$0.113\pm	0.009$&	$0.267\pm	0.01$&	$0.018\pm	0.002$&  $\bm{0.302\pm0.01}$\\
\hline
Random        &$0.159\pm	0.02$&	$0.266\pm	0.02$&	$0.020\pm	0.003$&	$0.354\pm	0.01$\\
\hline
Kcentres      &$0.159\pm	0.01$&	$0.267\pm	0.01$&	$0.017\pm	0.002$&	$0.380\pm	0.02$\\
\hline
GA (clust) sup&$0.112\pm	0.01$&	$0.259\pm	0.01$&	$0.016\pm	0.002$&	$0.350\pm	0.02$\\
\hline
FS            &$\bm{0.097\pm   0.008}$&  $0.260\pm	0.01$&	$0.017\pm	0.006$&	$0.339\pm	0.05$\\
\hline
GA (clust) MST&$0.143\pm	0.01$&	$\bm{0.254\pm	0.01}$&	$\bm{0.015\pm	0.002}$&	$0.378\pm	0.01$\\
\hline
GA MST        &$0.148\pm	0.01$&	$0.263\pm	0.01$&	$\bm{0.015\pm	0.002}$&	$0.402\pm	0.02$\\
\hline
\end{tabular}
\end{table*}\normalsize

\begin{figure}[!ht]
\centering
\subfigure[MNIST 1-NN]{%
\includegraphics[scale=0.35]{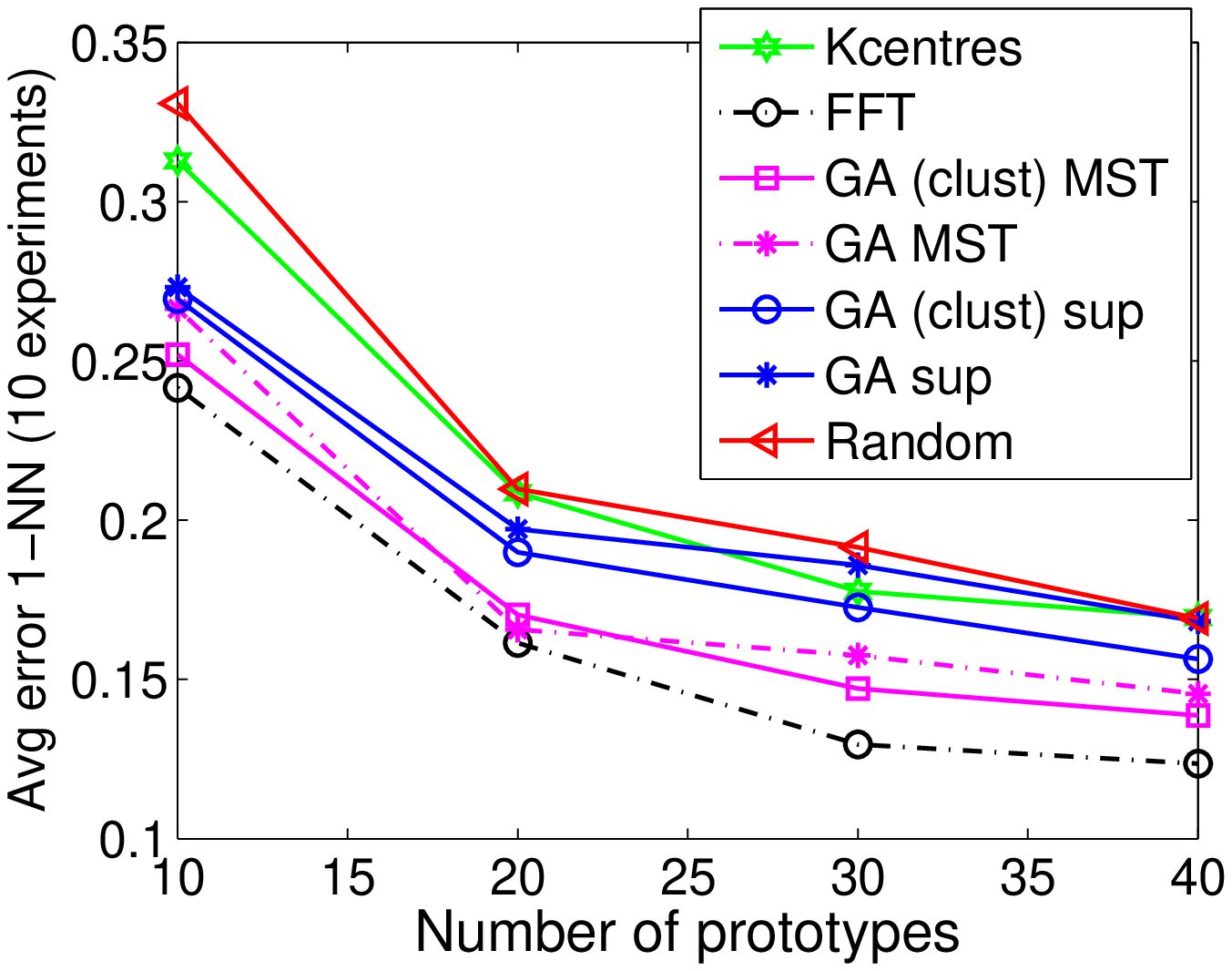}
\label{MNISTnn}%
}%
\subfigure[MNIST QDC]{
\includegraphics[scale=0.35]{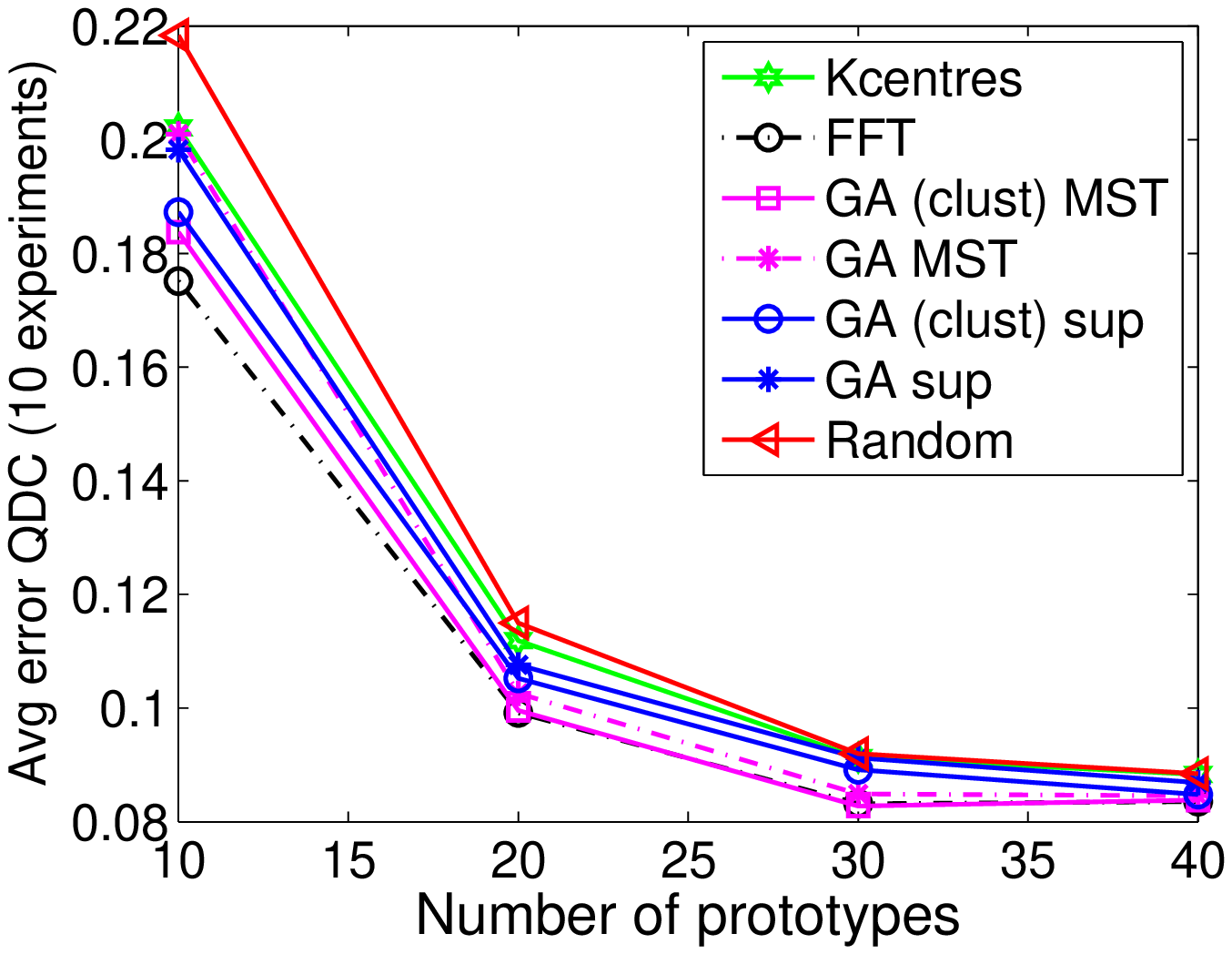}
\label{MNISTqdc}
}
\caption{Average errors for different numbers of prototypes in the MNIST data}
\label{LargedataresM}
\end{figure}

\begin{figure}[!ht]
\centering
\subfigure[SVHN 1-NN]{%
\includegraphics[scale=0.35]{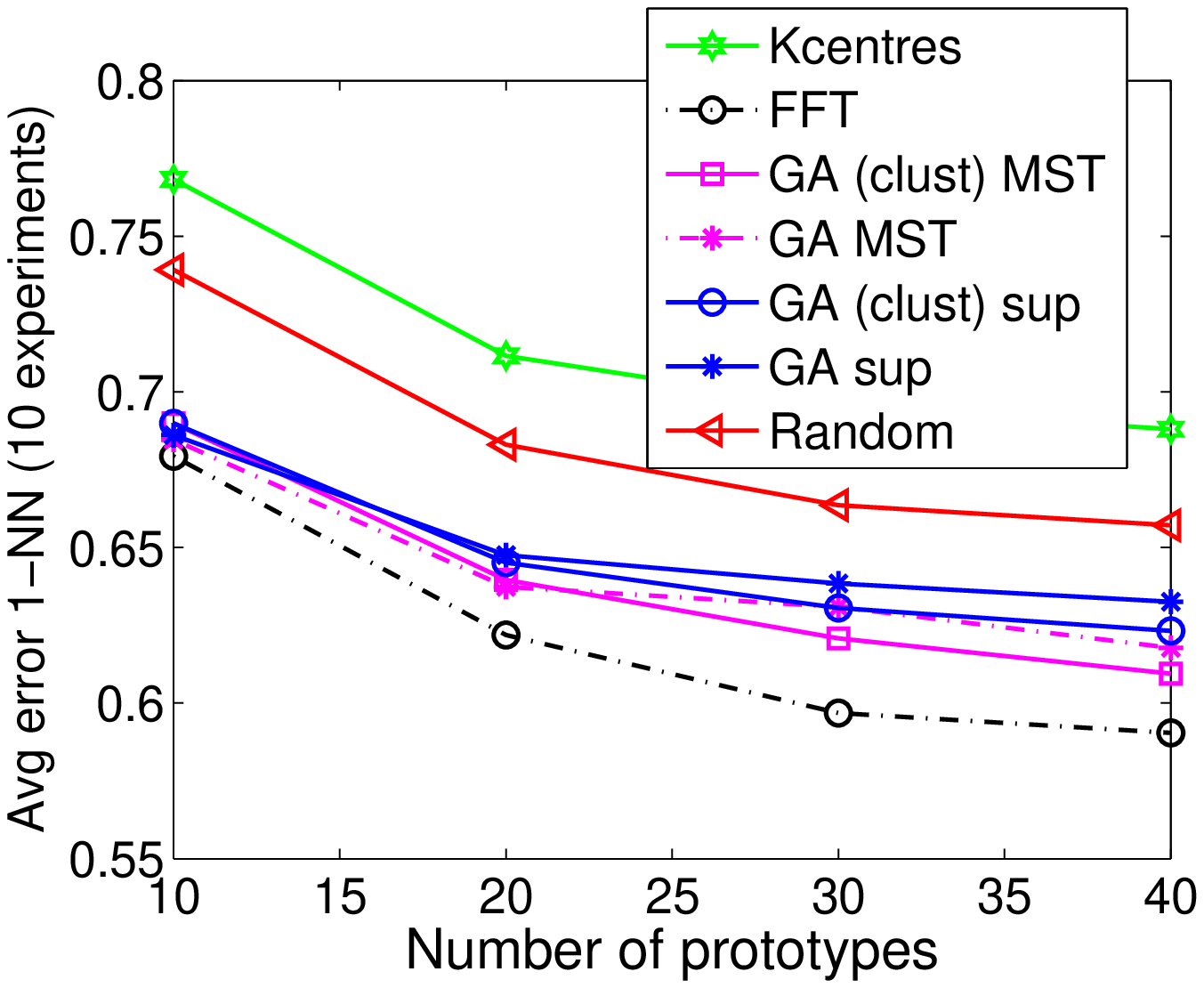}
\label{SVHNnn}%
}%
\subfigure[SVHN QDC]{
\includegraphics[scale=0.35]{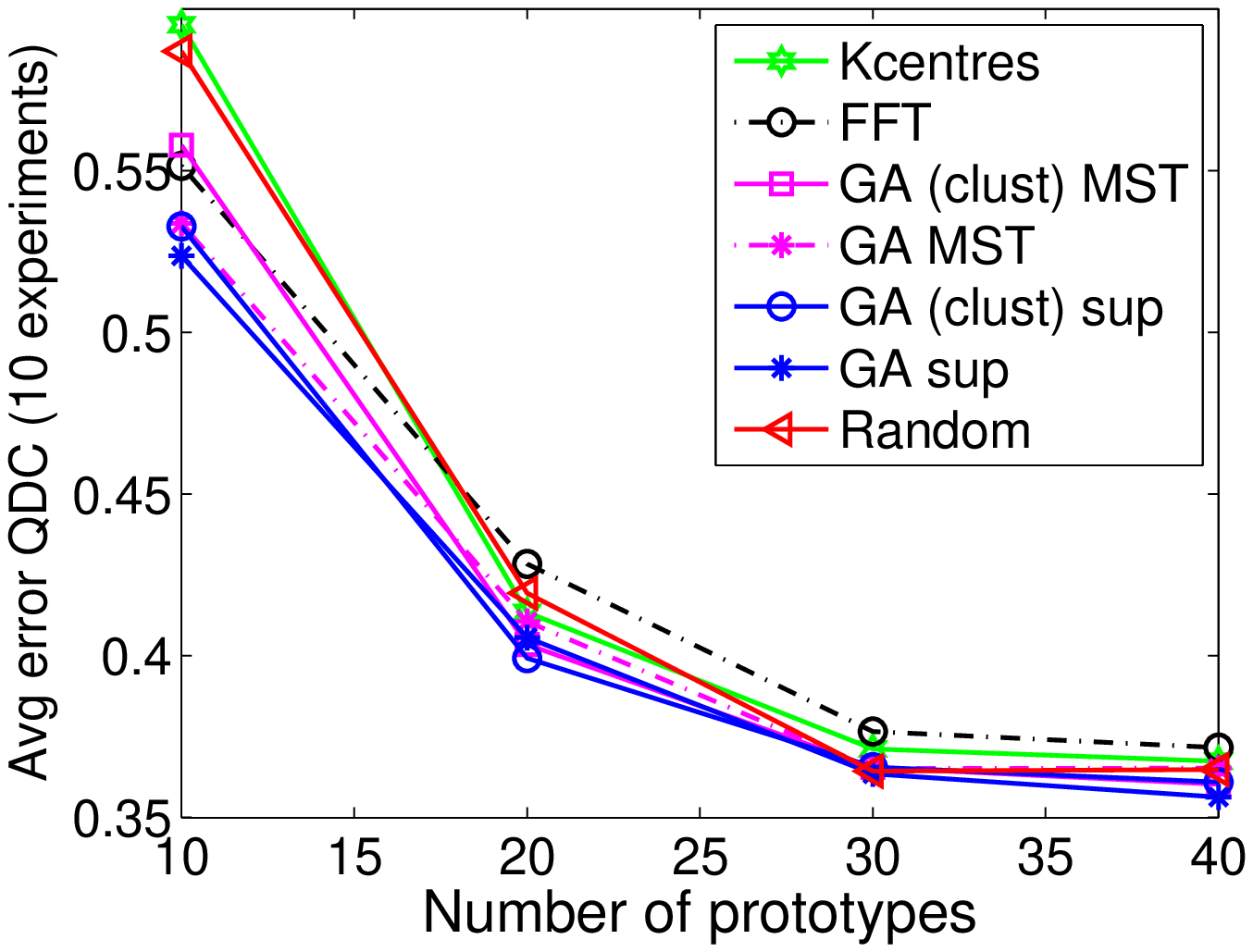}
\label{SVHNqdc}
}
\caption{Average errors for different numbers of prototypes in the SVHN data}
\label{LargedataresS}
\end{figure}

\begin{figure}[!ht]

\subfigure[Two prototypes]{
\includegraphics[scale=0.4]{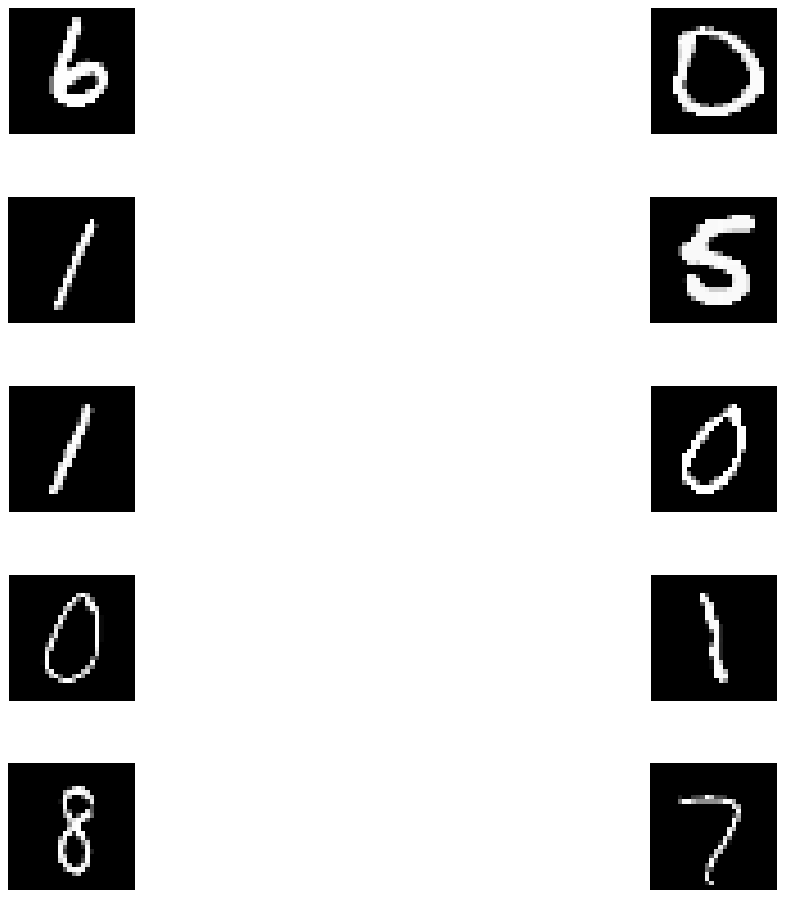}
\label{protMNIST2}%
}%
\subfigure[Ten prototypes]{
\includegraphics[scale=0.3]{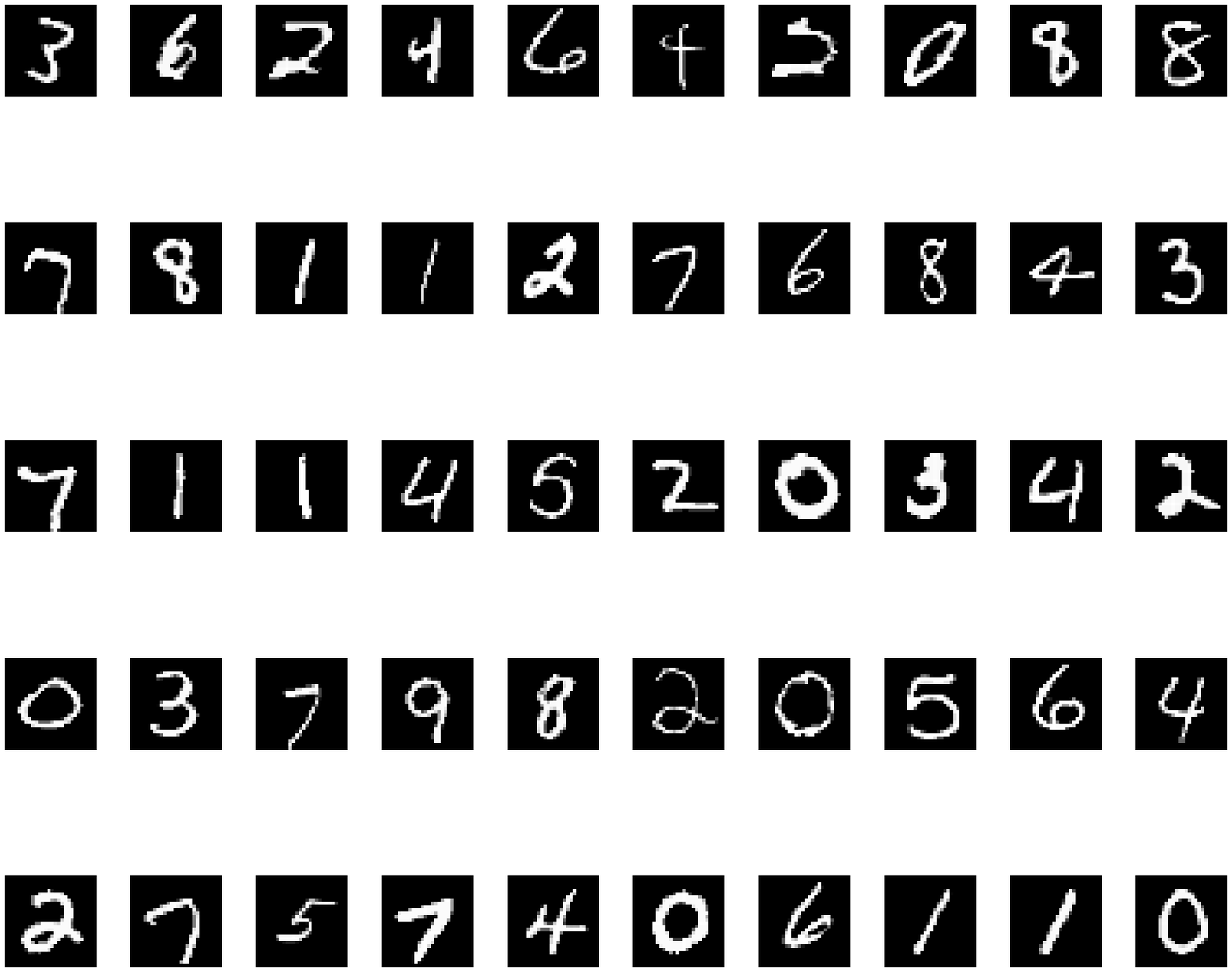}
\label{protMNIST10}
}
\caption{The prototypes selected by different methods on the MNIST dataset from top to bottom: GA (clust) MST, GA MST, GA (clust) sup, GA sup, random selection}
\label{protMNIST}
\end{figure}

\begin{table*} [!ht] \centering
\caption{Classification errors using a randomly selected training set of 12 358 objects for Youtube when selecting 10, 20 and 30 prototypes out of 247 170 objects set. Best results are in bold.}
\label{resYoutube1000}\small
\begin{tabular}{|c|c|c|c|}
\hline
\backslashbox{Method}{\ \# Prot.}&10&20&30\\
\hline
GA(clust)MST+QDC&\textbf{0.3431}&0.1831&0.1685\\
\hline
GA MST+QDC&0.3617&0.1936&0.1623 \\
\hline
GA(clust)sup+QDC&0.3926&0.1957&0.1586\\
\hline
GA sup+QDC&0.3494&0.1986&0.1619\\
\hline
random+QDC& 0.3551&0.1988&0.1634\\
\hline
FFT+QDC&  0.3963 & 0.2006&0.1614\\
\hhline{|=|=|=|=|}
GA(clust)MST+1-NN&0.3694&\textbf{0.1423}&0.1121\\
\hline
GA MST+1-NN&0.4022&0.1655&\textbf{0.1064}\\
\hline
GA(clust)sup+1-NN&0.4503&0.1668&0.1121\\
\hline
GA sup+1-NN&0.4032&0.1900&0.1219\\
\hline
random+1-NN& 0.4117&0.1927&0.1227\\
\hline
FFT+1-NN& 0.4275 & 0.1555&0.1108\\
\hline
\end{tabular}
\end{table*}\normalsize

\begin{table*} [!ht] \centering
\caption{Classification errors using a training set of 53113 objects and the standard test set of 26032 for SVHN2 when selecting 10, 20 and 30 prototypes out of the set of half a million objects. Best results are in bold.}
\label{resSVHNextra}
\begin{tabular}{|c|c|c|c|}
\hline
\backslashbox{Method}{\ \# Prot.}&10&20&30\\
\hline
GA(clust)MST+QDC&0.4678&0.3175&0.2404\\
\hline
GA MST+QDC&0.5043&0.3247 & 0.2470 \\
\hline
GA(clust)sup+QDC&0.4823&0.3090&0.2435\\
\hline
GA sup+QDC&0.4661&\textbf{0.3082}&\textbf{0.2402}\\
\hline
random+QDC& 0.5197& 0.3241&0.2480\\
\hline
FFT+QDC&  0.4801 & 0.3136&0.2414\\
\hhline{|=|=|=|=|}
GA(clust)MST+1-NN&0.5308&0.3925 &0.3582\\
\hline
GA MST+1-NN&0.5539& 0.4015 &0.3592\\
\hline
GA(clust)sup+1-NN&\textbf{0.4466}&0.3475&0.3224\\
\hline
GA sup+1-NN&0.4554&0.3742&0.3275\\
\hline
random+1-NN& 0.5831& 0.4209 &0.4018\\
\hline
FFT+1-NN& 0.5638 & 0.4141&0.3730\\
\hline
\end{tabular}
\end{table*}
\begin{figure*}[!ht]
\centering
\subfigure[Zongker]{%
\includegraphics[scale=0.3]{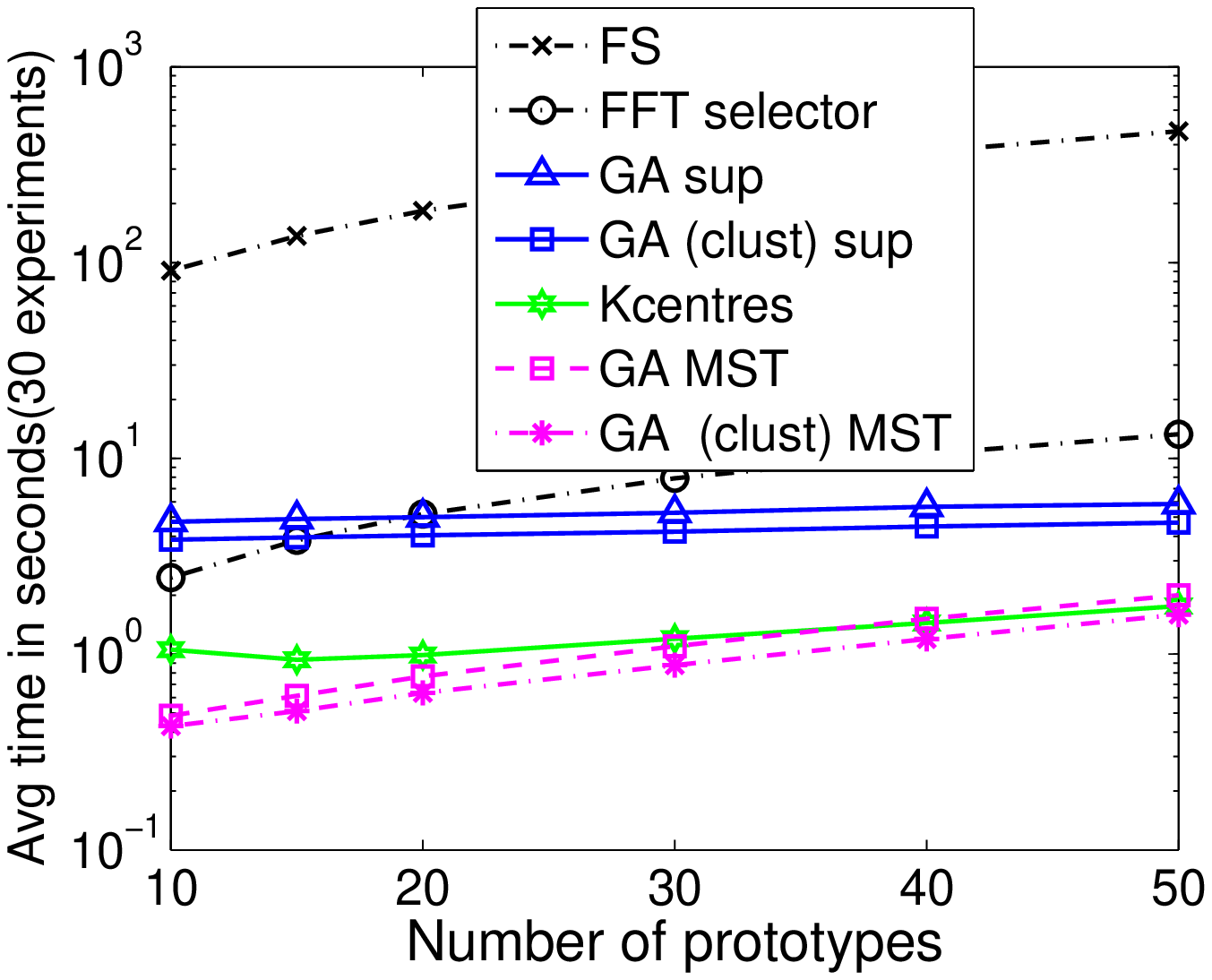}
\label{al}%
}%
\subfigure[Pendigits]{%
\includegraphics[scale=0.3]{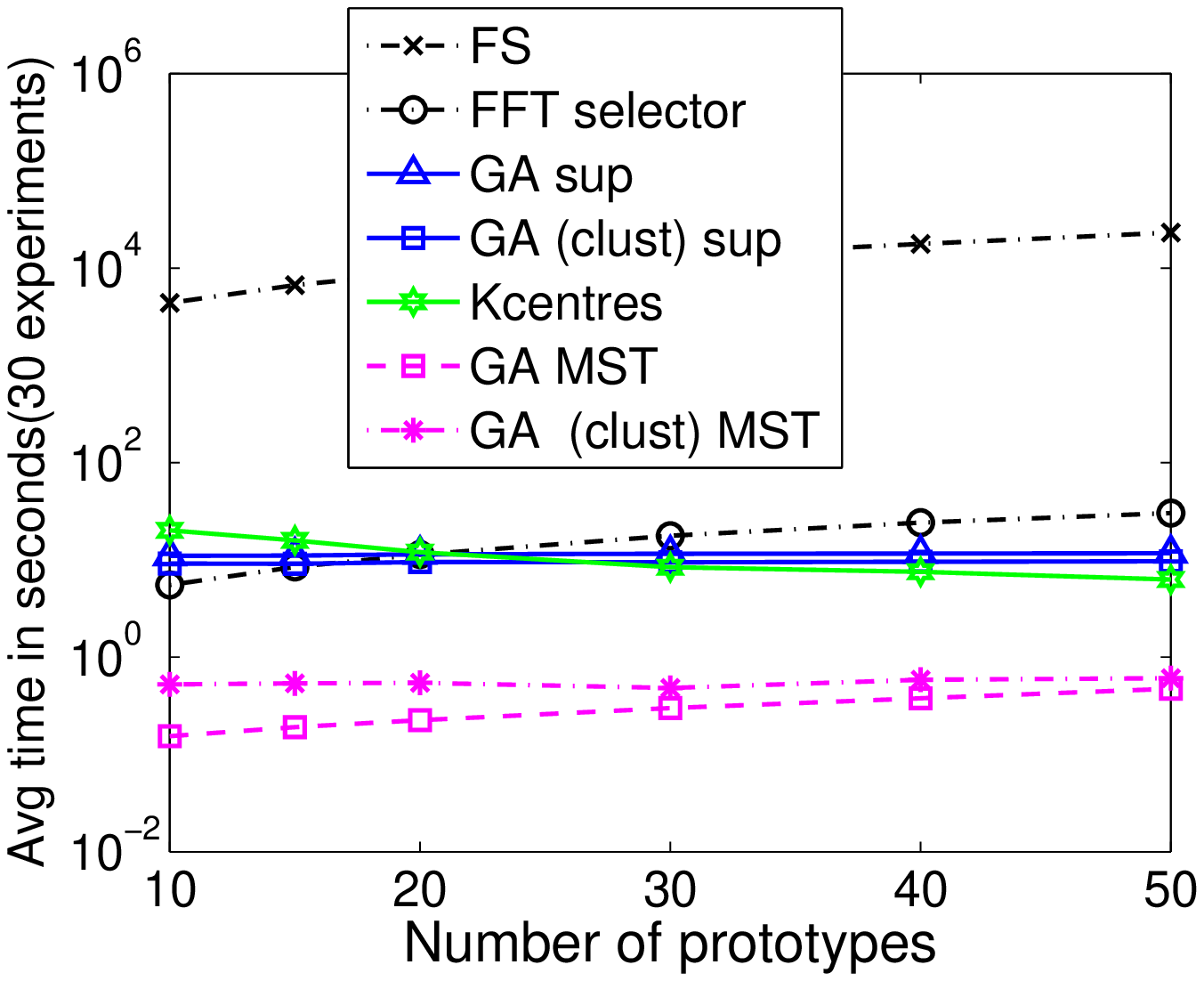}
\label{bl}%
}%
\subfigure[Diabetes]{
\includegraphics[scale=0.3]{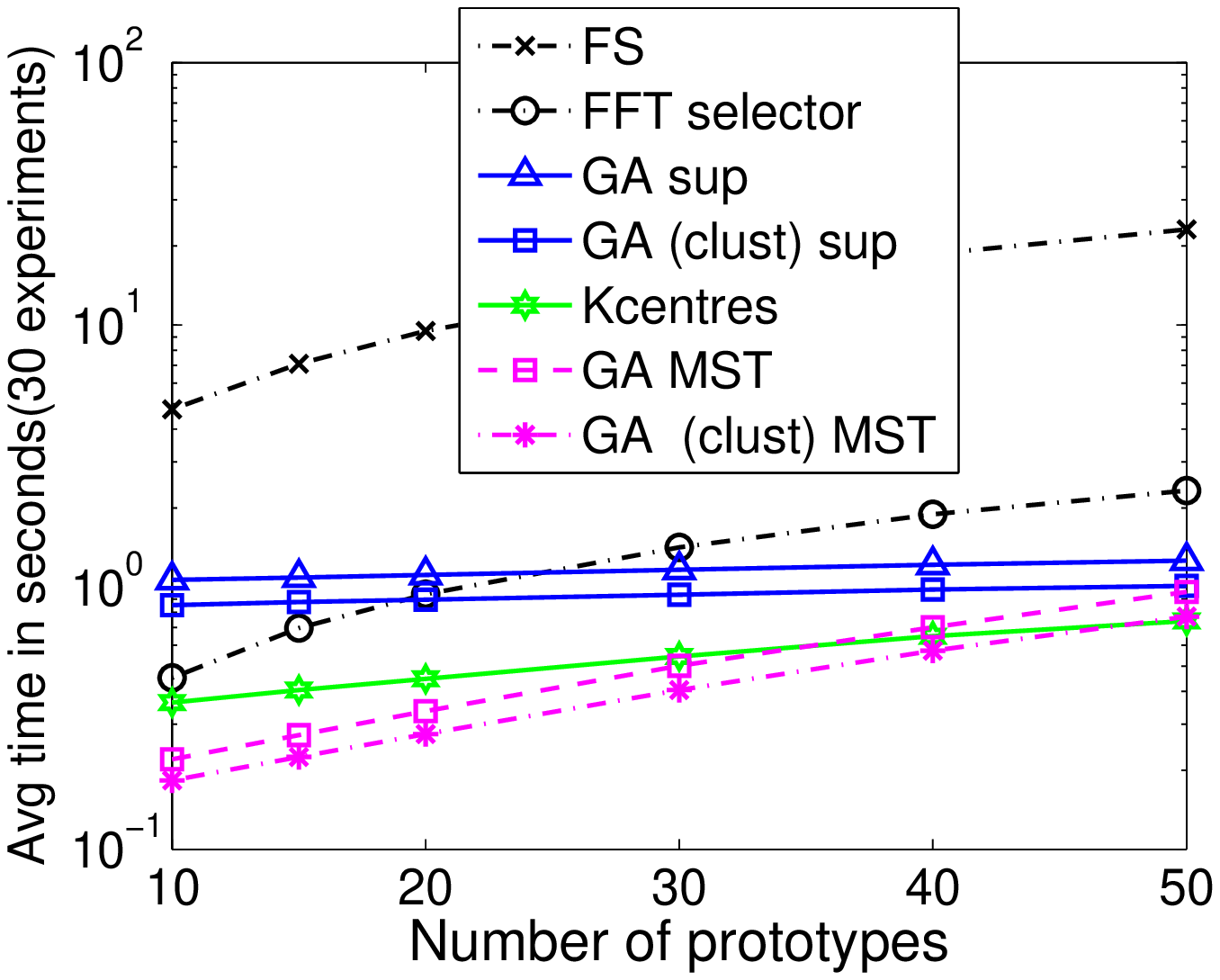}
\label{cl}
}
\subfigure[MNIST]{%
\includegraphics[scale=0.3]{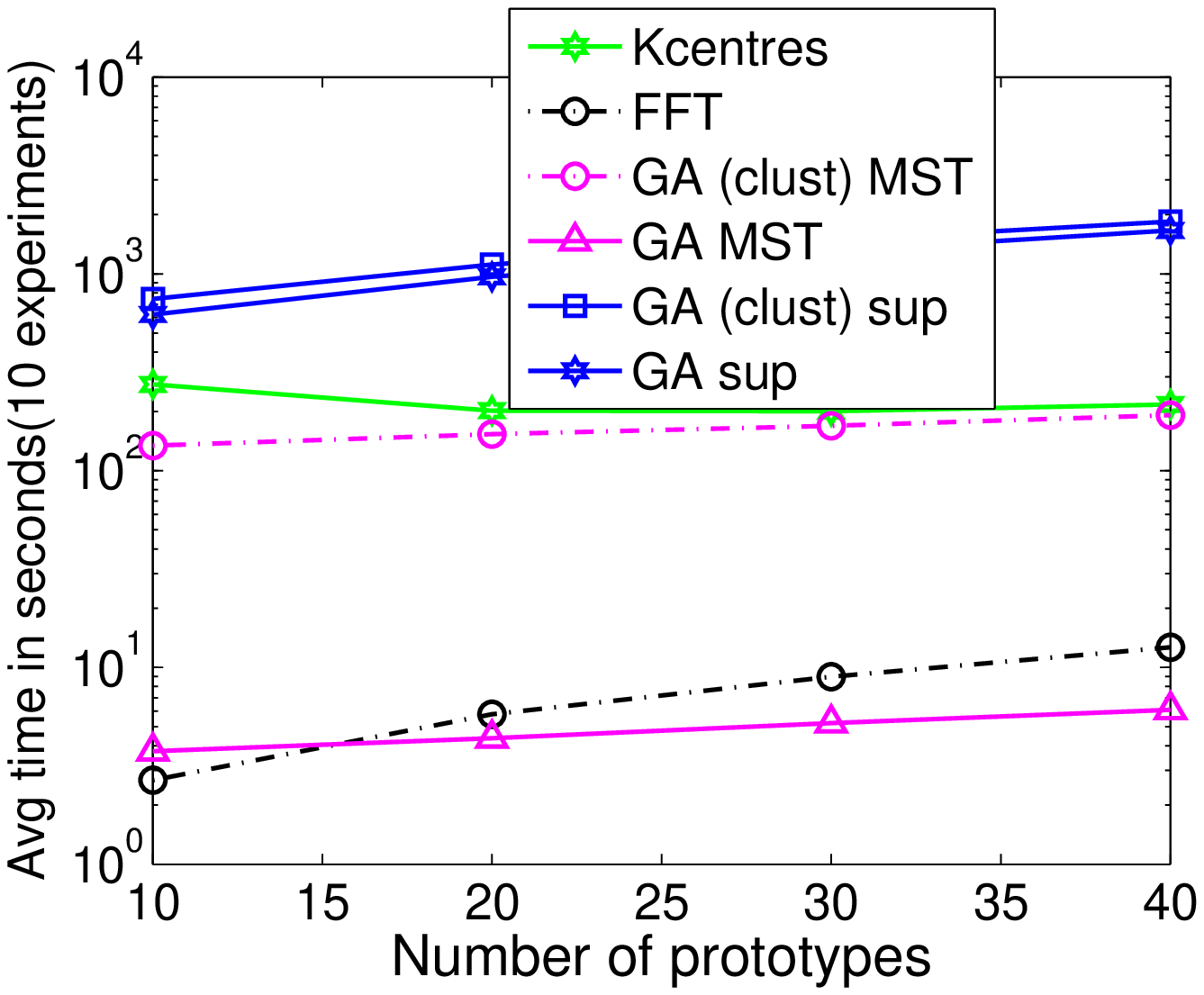}
\label{dl}%
}%
\subfigure[SVHN]{%
\includegraphics[scale=0.3]{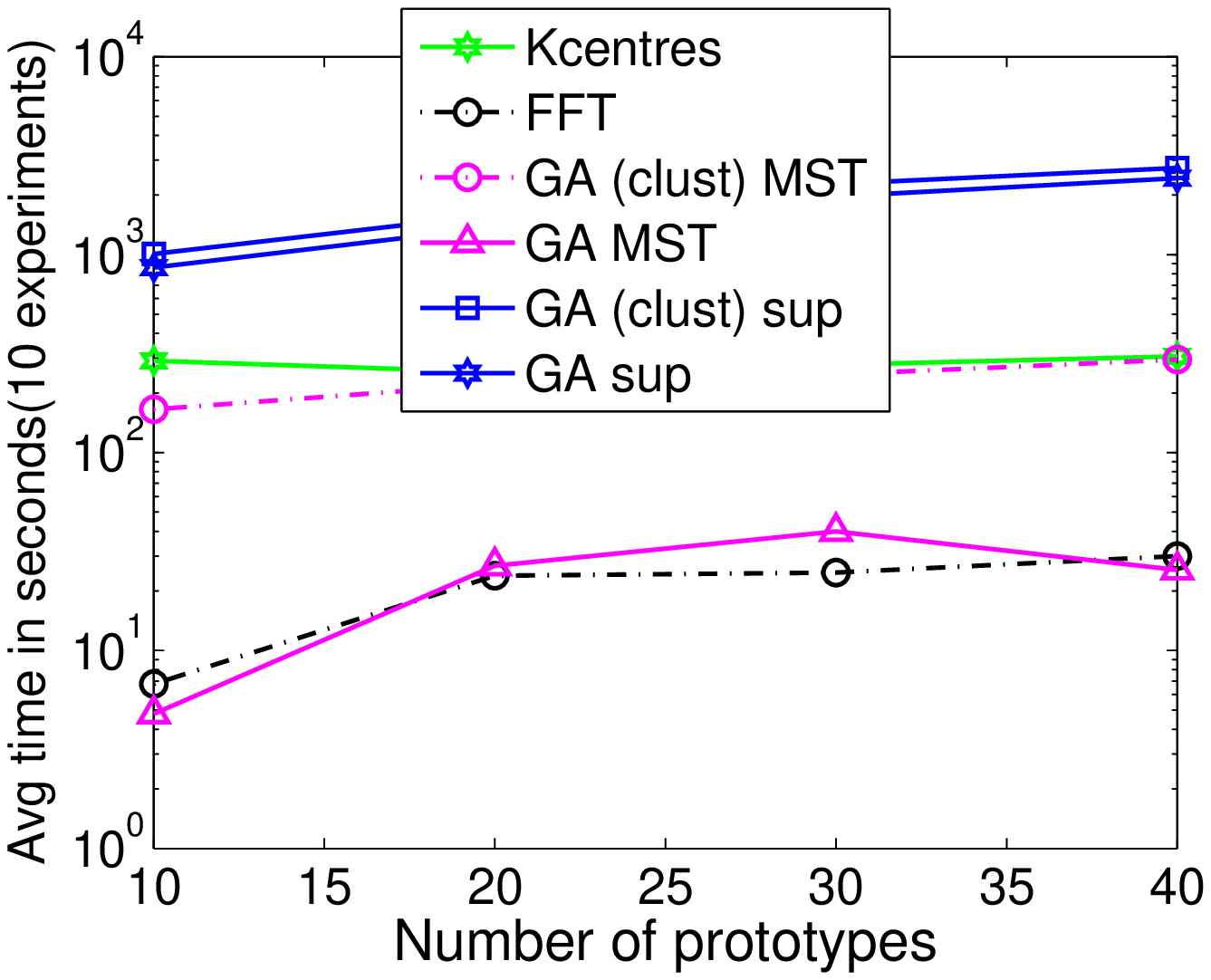}
\label{el}%
}%
\caption{Average times in seconds plotted in log scale}
\label{times}
\end{figure*}

\begin{figure*}[!ht]
\centering
\subfigure[Zongker]{%
\includegraphics[scale=0.20]{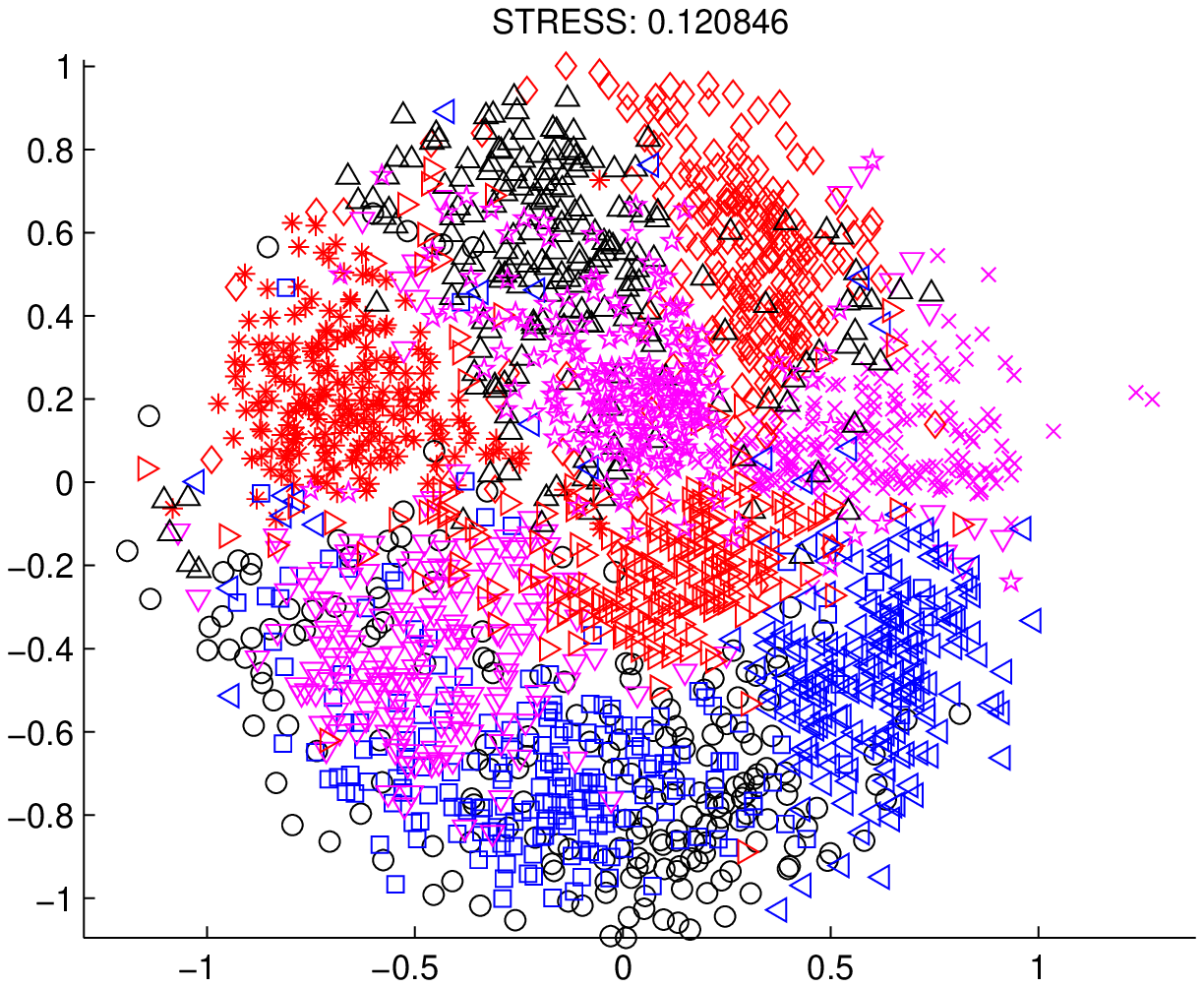}
\label{mdsa}%
}%
\subfigure[Pendigits]{%
\includegraphics[scale=0.15]{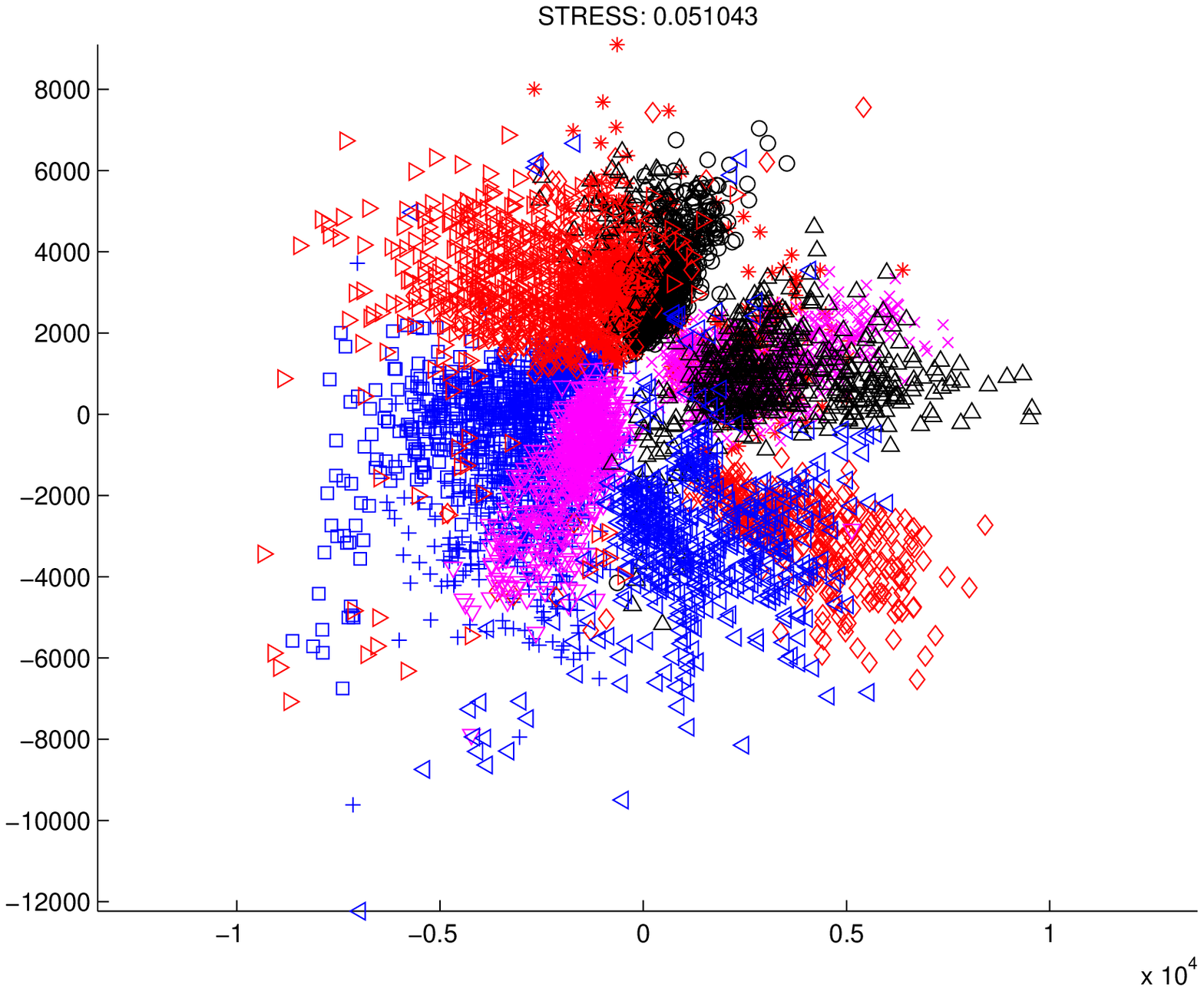}
\label{mdsb}%
}%
\subfigure[Diabetes]{%
\includegraphics[scale=0.20]{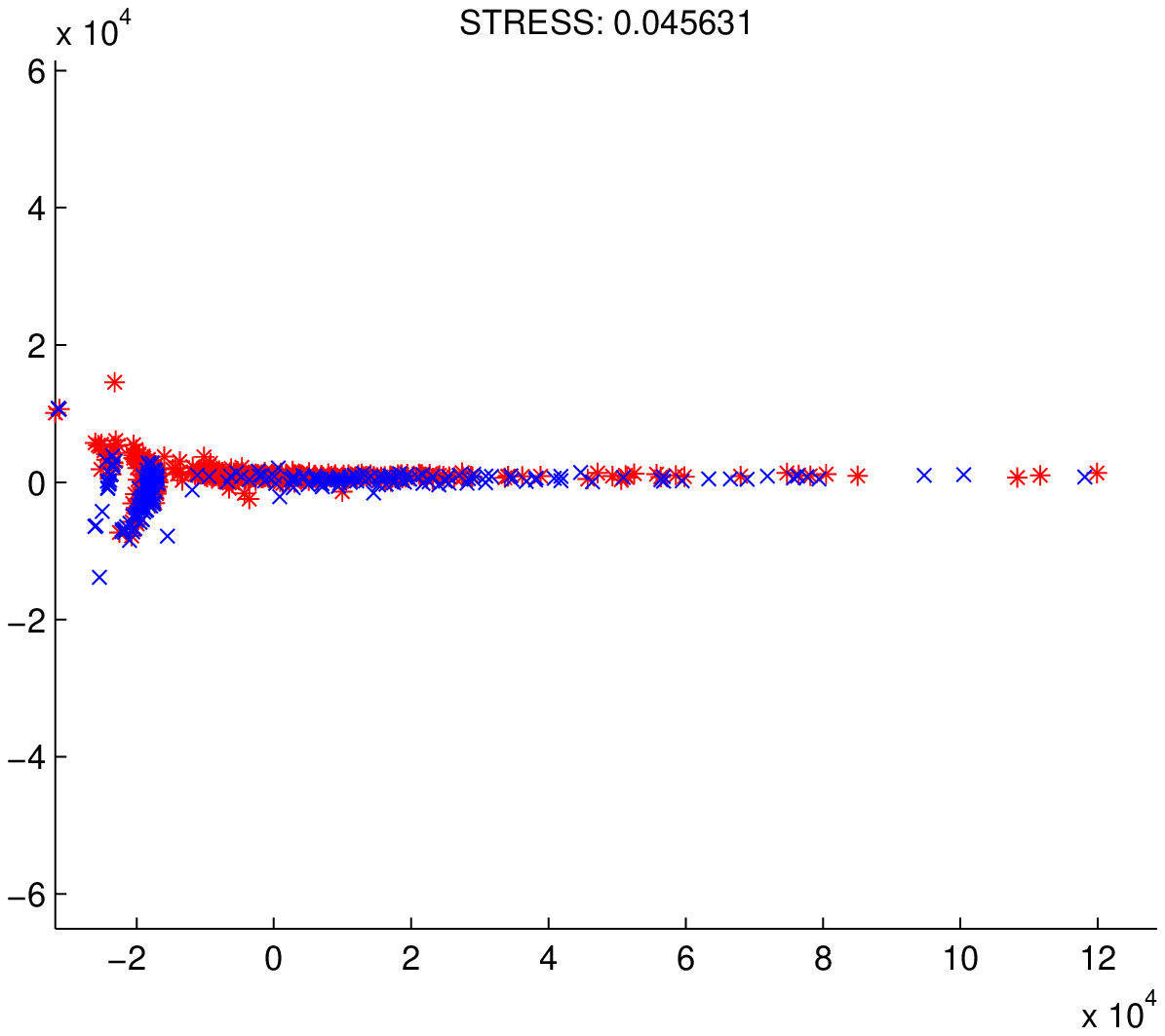}
\label{mdsc}%
}%
\subfigure[XM2VTS]{
\includegraphics[scale=0.20]{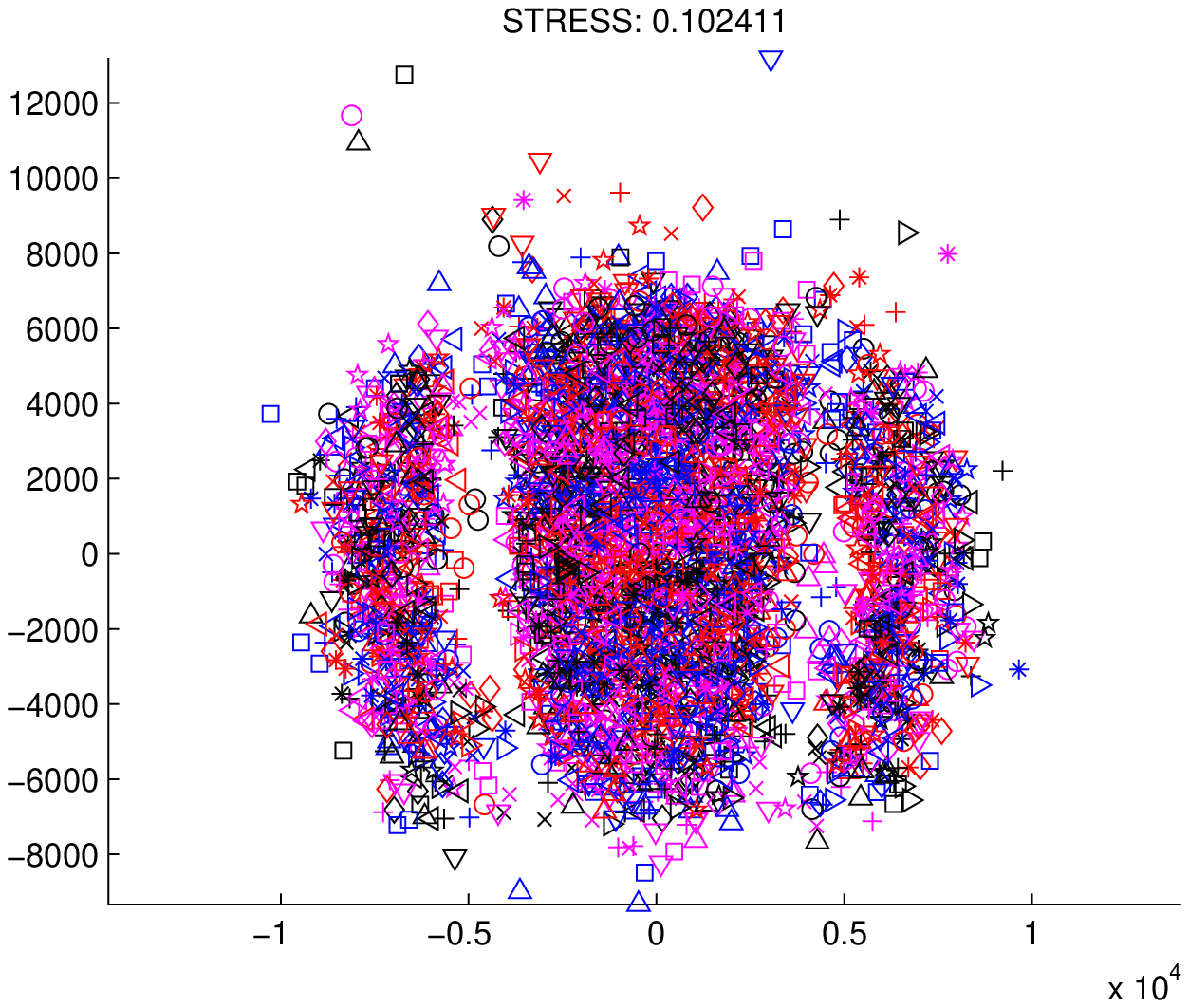}
\label{mdsd}
}
\subfigure[MNIST]{%
\includegraphics[scale=0.20]{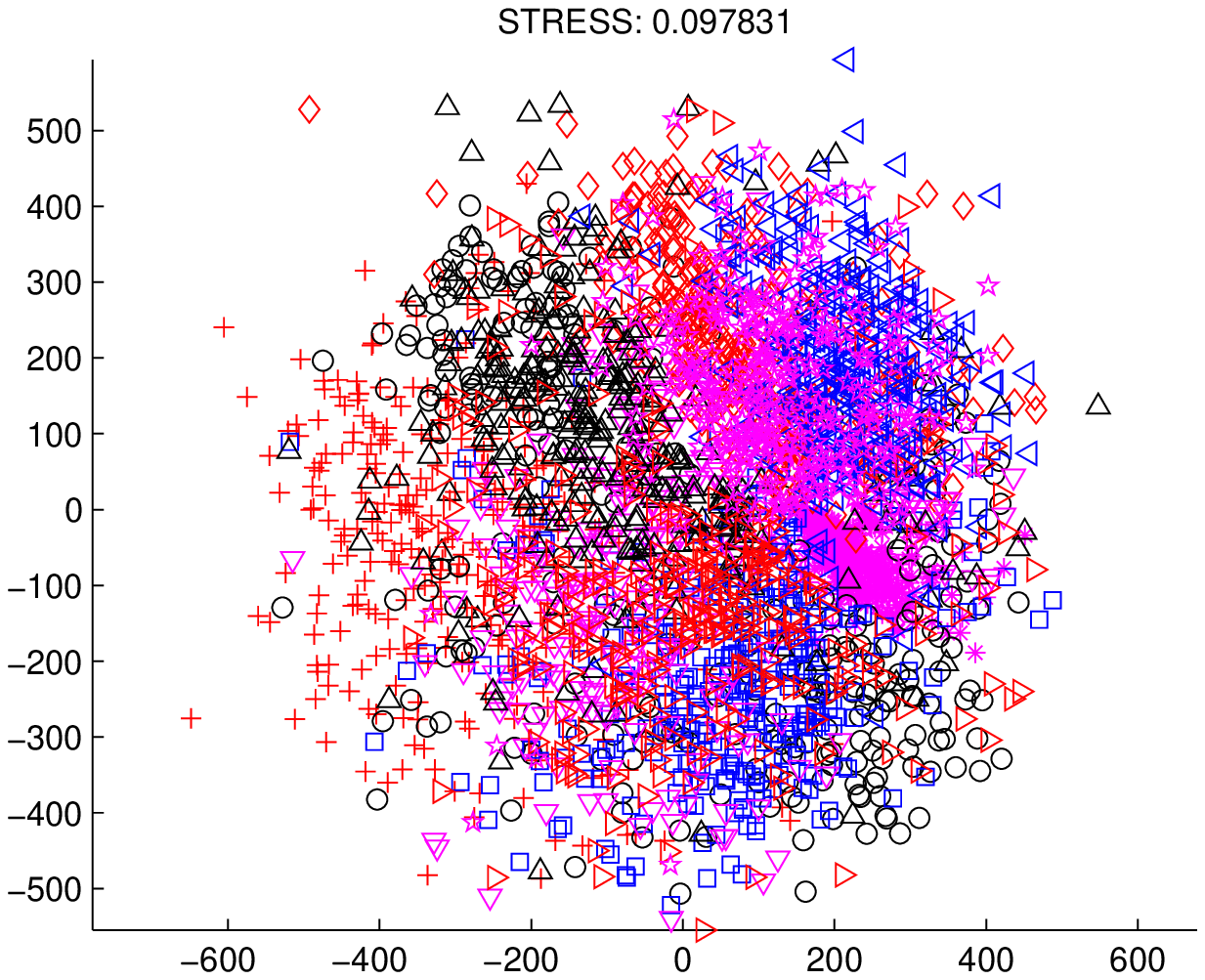}
\label{mdse}%
}%
\subfigure[SVHN]{%
\includegraphics[scale=0.20]{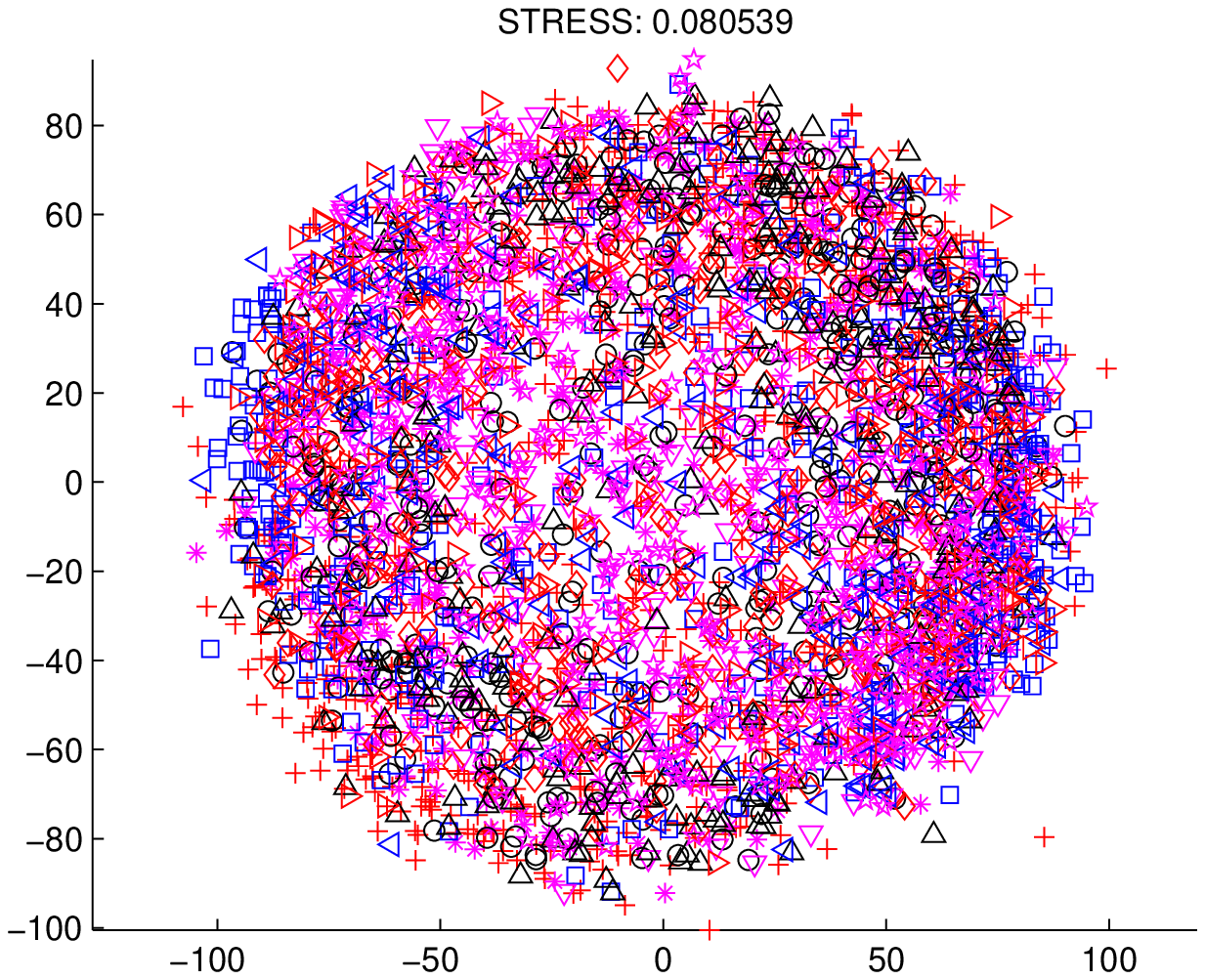}
\label{mdsf}%
}%
\subfigure[YouTube]{%
\includegraphics[scale=0.20]{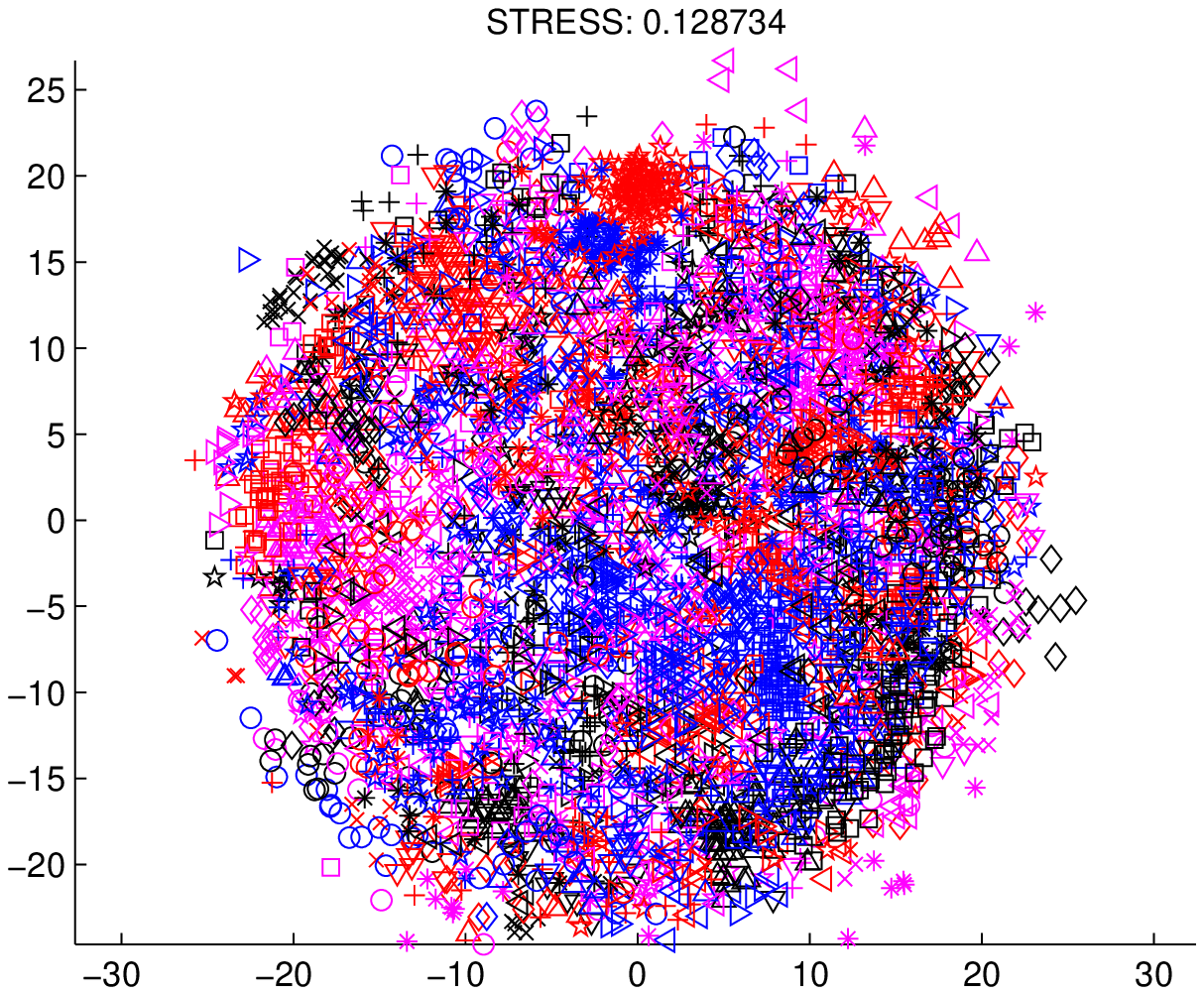}
\label{mdsg}%
}%
\caption{MDS mappings}
\label{mds}
\end{figure*}
\indent Table~\ref{resultssmall} presents the average errors over 30 experiments for 20 prototypes for small datasets where dissimilarities are already given. In the case of the MNIST and SVHN datasets the dissimilarities must be computed on demand, Figs.~\ref{LargedataresM} and ~\ref{LargedataresS} show the results averaged over 10 experiments for 10, 20 30 and 40 prototypes. \\
\indent It can be seen in Table~\ref{resultssmall} that, for 20 prototypes, the proposed unsupervised GA with clustering is the best or among the best in two of the datasets, Diabetes and Pendigits, while the supervised version outperform the other procedures in the XM2VTS dataset. Only in the Zongker dataset the proposed methods where not as successful as the best method, however the second best method was the supervised GA. These results show that the proposed GAs are good for prototype selection. In the next set of experiments, we will show that they are able to select prototypes out of very large datasets, without having to resort to the straightforward solution of randomly sampling the datasets.\\
\indent In the experiments on the large-scale datasets we can see from Figs.~\ref{LargedataresM} and ~\ref{LargedataresS} that the 1-NN benefits more from a systematic selection of the prototypes than the QDC. This can be seen from the improvements found when using the proposed systematic methods for selection compared to the random selection. By performing a $t$-test we detected that the improvements were statistically significant for the 1-NN in the MNIST and SVHN datasets. Our methods are almost always the best performing ones for the QDC classifier. However, the FFT is better than our procedures when using the 1-NN.\\
\indent Figures~\ref{protMNIST2} and~\ref{protMNIST10} show the digits returned as prototypes for MNIST when the methods select the best two and ten. It can be seen that the GAs returns digits that differ from each other, manifesting the suitability of the used criteria to allow diversity in shapes of the set of prototypes. It is clear that the random selection is not able to achieve this diversity. The supervised GA returns perfectly handwritten digits, which can be expected since those digits are the ones that represent their classes better. Therefore, depending on the problem, the selected prototypes may provide interesting information for other analysis different from classification. From the results on Youtube data in Table~\ref{resYoutube1000}, it is interesting to see that with only 30 prototypes we obtain remarkably good results for a 1045-class problem. This means that we do not need objects from all the classes to generate a good DS. We use all the classes for training but only 10, 20 or 30 objects are used, in total, to create the space where the training set is mapped. The GA maximizing the MST outperformed the other methods in all cases.\\
Table~\ref{resSVHNextra} shows the classification errors when selecting the prototypes out of the set of half a million images. It can be seen that the supervised selection by GA provides the best results. Note that it uses the complete set of half a million objects to compute the selection criterion in each GA iteration, in contrast to the unsupervised version that only needs the dissimilarities among the prototypes being evaluated to compute the criterion. Thereby, it is useful to use as many objects as possible to compute the selection criterion for this difficult dataset. Since the computation of the supervised criterion is linear in the number of objects, the computation cost is affordable.\\
\indent The computation times for some of the datasets reported in Fig.~\ref{times} show that the GA with unsupervised criterion is the fastest method in all the datasets, except for 10 prototypes in MNIST and SVHN1. The supervised proposal is comparable to other unsupervised methods. This analysis, together with the computational complexity analysis, indicates that the proposed GAs are able to scale well to large datasets when the final goal is the selection of a small set of prototypes. However, the fitness function must also be designed scalable. In addition, GAs are embarrassingly parallelizable, this means that they can be easily decomposed into parallel subtasks and therefore they can be further optimized for scalability. \\
\indent We analyzed relations between performance of methods and data distribution by inspecting the MDS plots. We found that the supervised method handles well datasets with homogeneous distributions or with class overlap as the one in Fig.~\ref{mdsa}. In contrast, the MST-based unsupervised criterion copes better with non-homogeneous distributions where we can find, inside the same class, densely populated regions as well as sparse ones as in Fig.~\ref{mdsb}. In addition, the MST-based GA handles well easier problems and elongated classes as in the Diabetes dataset in Fig.~\ref{mdsc}. We found that among the competitors the FFT provided reasonably good results especially for easier problems and when using the 1-NN classifier. The FFT might be preferred in case of clear separable classes and the GA, especially the supervised version, for more difficult problems with overlapping classes. This can be expected since the GA refines the set of prototypes en each iteration while the FFT adds one prototype in each iteration and the method stops when the desired cardinality of the set is reached without any refinement of this set.\\
\begin{figure*}[!ht]
\centering
\subfigure[Accuracy results]{%
\includegraphics[scale=0.4]{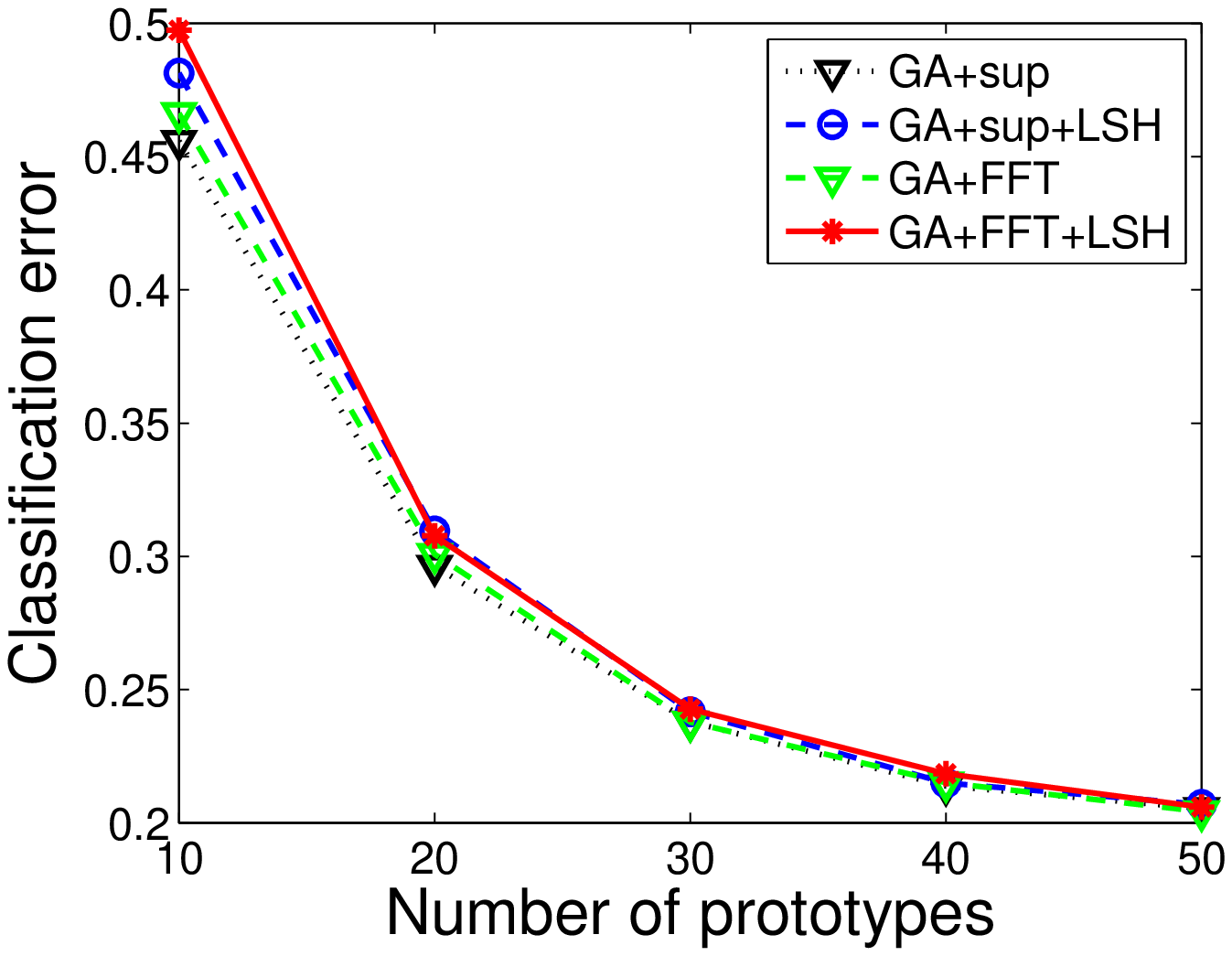}
\label{sh1}%
}%
\subfigure[Execution time results]{%
\includegraphics[scale=0.4]{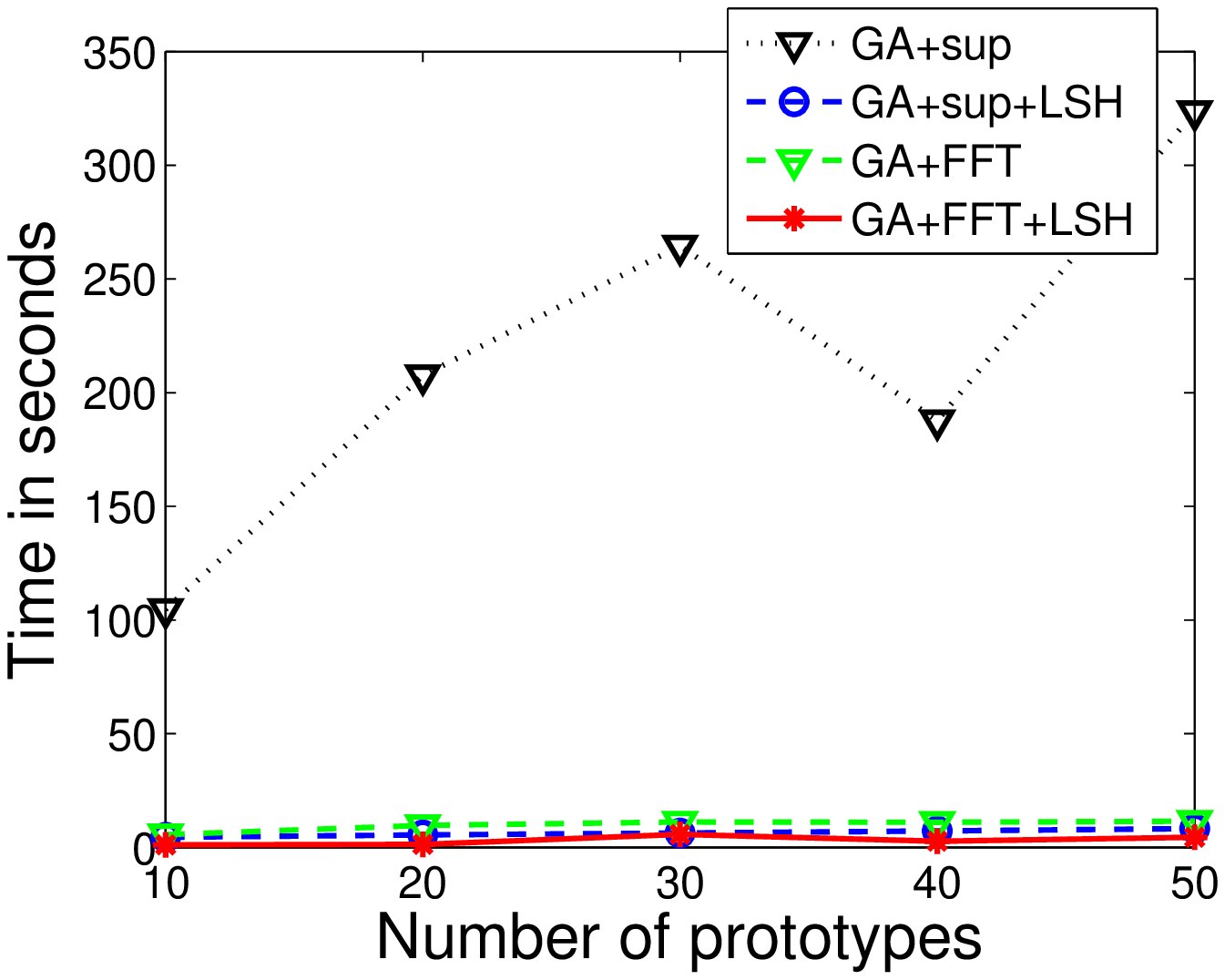}
\label{sh2}%
}%
\caption{Accuracy and execution times results when using LSH to speed up the proposed prototype selection methods in SVHN dataset}
\label{shSVHN}
\end{figure*}
\begin{figure*}[!ht]
\centering
\subfigure[Accuracy results]{%
\includegraphics[scale=0.4]{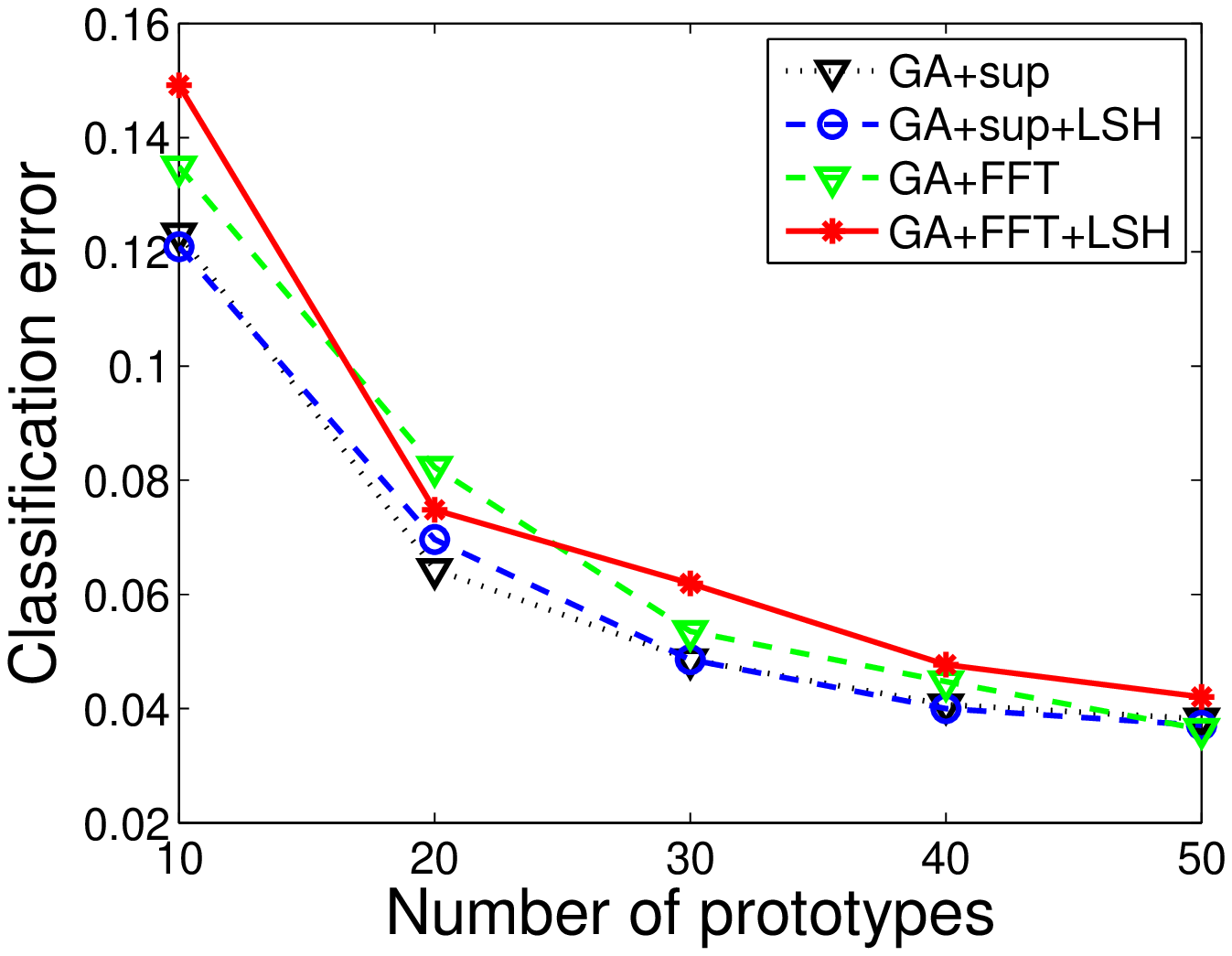}
\label{sh1}%
}%
\subfigure[Execution time results]{%
\includegraphics[scale=0.4]{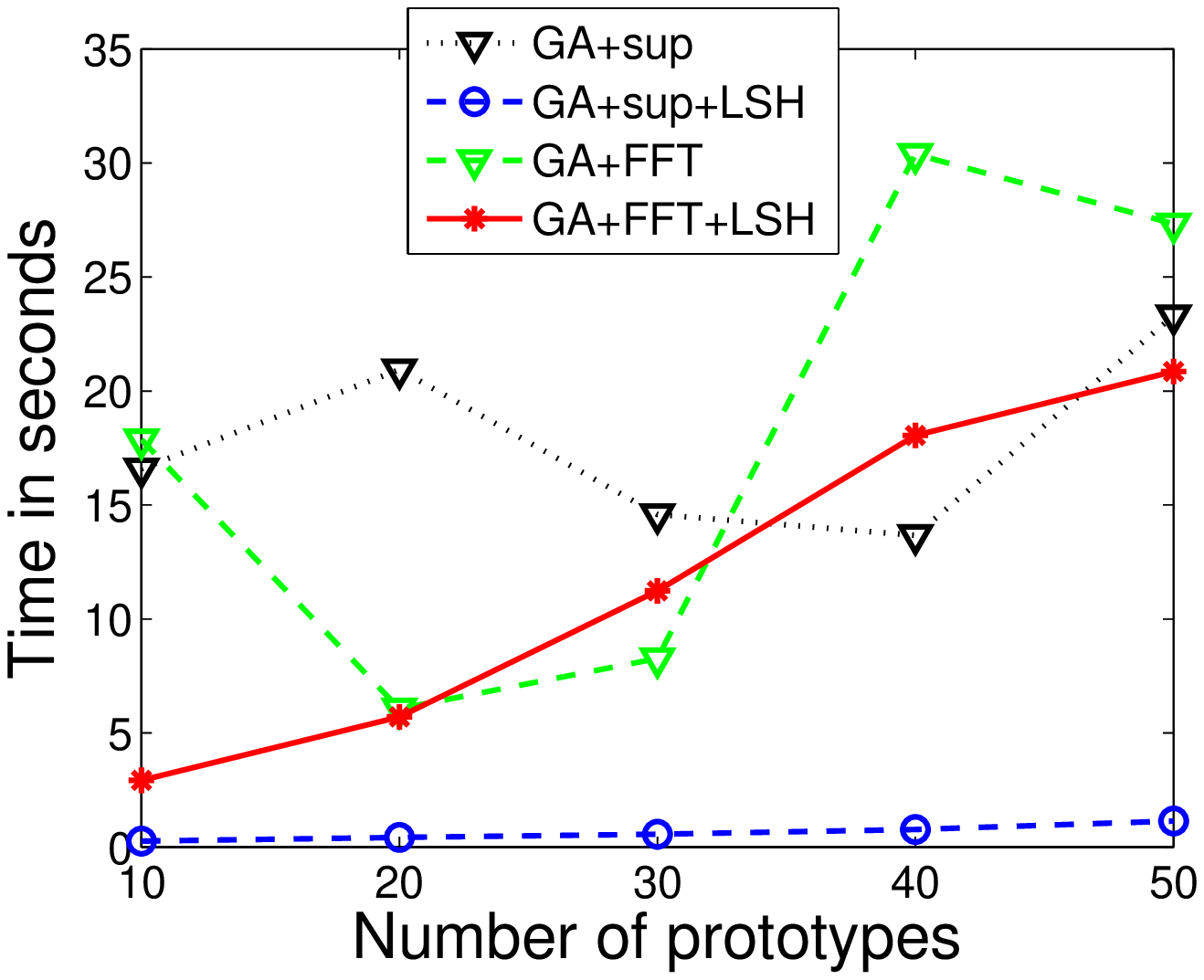}
\label{sh2}%
}%
\caption{Accuracy and execution times results when using SH to speed up the proposed prototype selection methods in XM2VTS dataset}
\label{shXM2VTS}
\end{figure*}
Figs~\ref{shSVHN}-\ref{shXM2VTS} show the results for the SVHN and XM2VTS datasets when using spherical hashing to speed-up the proposed methods, Here we name this version LSH, from Limited Spherical Hashing since we do not use the original procedure. Instead, we use a limited version which uses the dissimilarities instead of the binary codification since the latter increases the classification error, while the use of dissimilarities still provides sufficient speed. We use 64 pivots to maintain low computing times. It can be seen that the LSH is able to decrease running times significantly, especially for the supervised criterion, without a significant decrease in accuracy. For the smaller dataset XM2VTS the increase in speed is not as significant as in the larger dataset SVHN.  
\section{Conclusions}
\label{sec:concl}
The selection of prototypes is a crucial step for classification in the dissimilarity space. In this paper we proposed two different GA-based scalable prototype selection methods. Our proposals achieve scalability by exploiting the suitability of genetic algorithms to find good trade-offs between time complexity and accuracy of the solution for our problem, by maintaining low asymptotic complexities in the two different fitness functions proposed, and by using dissimilarity-based hashing. Experimental results showed the validity of the proposals for selecting good prototypes and the runtime analysis showed that the methods are able to scale to large datasets. Other general approaches to cope with scalability include parallelism, stochastic methods, among others. Our proposals are able to obtain compact and discriminative dissimilarity representations which can be used for classification. \\
\indent The proposed unsupervised method is the fastest one since the evaluation of its criterion does not depend on the size of the dataset but on the number of prototypes. However by using hashing the proposed supervised selection becomes as fast as the unsupervised one. After comparing the unsupervised and supervised methods for the selection, a question arises: is the label of a prototype really relevant? Another object with the same dissimilarities to other objects but a different label will likely generate the same result. However, the use of labels, in general, allow us to emphasize that we search for different objects, improving the coverage over the space of objects. In addition, in our procedure, we also ask that each of these prototypes must represent its class as good as possible. These requirements make the procedure especially good for datasets with overlap among the classes. However, in other cases, unsupervised procedures may do equally well. Especially, if we are aiming at many more, or much less prototypes than classes, their labels will not help us.\\
\indent We found that it is profitable in some cases to inspect a large number of objects in the selection criterion as we do in our supervised proposal. This holds especially for difficult datasets since we find improvements over the other methods when using this strategy. In addition, the computational burden when including even millions of objects in the criterion computation is affordable, as the supervised criterion is linear in the number of objects. For not very complicated problems, the unsupervised selection that only uses the distances among the prototypes in the criterion computation is sufficiently good, and no further improvements are found by involving the large datasets in the criterion computation.\\

\section*{Acknowledgement}
We acknowledge financial support from the FET programme within the EU FP7, under the SIMBAD project (contract 213250).

\bibliographystyle{elsarticle-num}
\bibliography{references}   
\end{document}